\documentclass[article,onefignum,onetabnum]{siamonline171218}

\usepackage{lipsum}
\usepackage{amsfonts}
\usepackage{graphicx}
\usepackage{epstopdf}
\usepackage{algorithmic}
\ifpdf
  \DeclareGraphicsExtensions{.eps,.pdf,.png,.jpg}
\else
  \DeclareGraphicsExtensions{.eps}
\fi

\usepackage{enumitem}
\setlist[enumerate]{leftmargin=.5in}
\setlist[itemize]{leftmargin=.5in}

\headers{}{Huang and Pesenti}

\title{Marginal Fairness: Fair Decision-Making under Risk Measures\thanks{This version: \today.
}}

\author{Fei Huang
\thanks{UNSW Business School, School of Risk and Actuarial Studies, Sydney, New South Wales, Australia
  (\email{feihuang@unsw.edu.au}).}
\and Silvana M. Pesenti\thanks{Department of Statistical Sciences, University of Toronto, Canada
  (\email{silvana.pesenti@utoronto.ca}).}
}

\usepackage{amsopn}

\makeatletter
\newcommand*{\addFileDependency}[1]{
  \typeout{(#1)}
  \@addtofilelist{#1}
  \IfFileExists{#1}{}{\typeout{No file #1.}}
}
\makeatother

\usepackage{todonotes}
\usepackage{float}
\usepackage{setspace}
\usepackage{amsmath}
\usepackage{enumitem}
\usepackage{amssymb,amsmath}
\usepackage{wrapfig}
\usepackage{dsfont, mathrsfs}
\usepackage{tikz}
\usetikzlibrary{arrows.meta, positioning, calc}
\usepackage{tabularx}
\usepackage{booktabs}

\newsiamremark{assumption}{Assumption}
\newsiamremark{remark}{Remark}
\newsiamremark{hypothesis}{Hypothesis}
\crefname{hypothesis}{Hypothesis}{Hypotheses}
\newsiamthm{example}{Example}
\newsiamremark{optimisation}{Optimisation}

\newcommand{\R}{{\mathds{R}}}
\newcommand{\N}{{\mathds{N}}}
\newcommand{\Id}{{\mathds{1}}}

\newcommand{\E}{{\mathbb{E}}}
\renewcommand{\L}{{\mathbb{L}^2}}
\renewcommand{\P}{{\mathbb{P}}}

\renewcommand{\t}{{\mathbf{t}}}
\renewcommand{\d}{{\mathbf{d}}}
\newcommand{\D}{{\mathbf{D}}}
\newcommand{\x}{{\mathbf{x}}}
\newcommand{\X}{{\mathbf{X}}}
\newcommand{\z}{{\mathbf{z}}}
\newcommand{\Z}{{\mathbf{Z}}}
\newcommand{\V}{{\mathbf{V}}}
\renewcommand{\v}{{\mathbf{v}}}
\newcommand{\W}{{\mathbf{W}}}

\newcommand{\F}{{\mathcal{F}}}

\newcommand{\mcH}{{\mathcal{H}}}
\newcommand{\mcK}{{\mathcal{K}}}
\newcommand{\mcJ}{{\mathcal{J}}}

\newcommand{\mfg}{{\mathfrak{g}}}

\newcommand{\VaR}{{\textrm{VaR}}}
\newcommand{\ES}{{\textrm{ES}}}

\newcommand{\supp}{{\textrm{supp}}}

\renewcommand{\Finv}{{\breve{F}}}
\newcommand{\diff}{{\rm d}}

\DeclareMathOperator*{\argmin}{arg\,min}

\newcommand{\ep}{{\varepsilon}}
\newcommand{\bPsi}{{\boldsymbol{\Psi}}}

\allowdisplaybreaks

\hypersetup{
  colorlinks   = true, 
  urlcolor     = blue, 
  linkcolor    = red, 
  citecolor   = blue 
}

\begin{document}

\maketitle

\begin{abstract}
This paper introduces \textit{marginal fairness}, a new individual fairness notion for equitable decision-making in the presence of protected attributes such as gender, race, and religion. This criterion ensures that decisions—based on generalized distortion risk measures—are insensitive to distributional perturbations in protected attributes, regardless of whether these attributes are continuous, discrete, categorical, univariate, or multivariate. 
To operationalize this notion and reflect real-world regulatory environments (such as the EU gender-neutral pricing regulation),  we model business decision-making in highly regulated industries (such as insurance and finance) as a two-step process:  
(i) a predictive modeling stage, in which a prediction function for the target variable (e.g., insurance losses) is estimated based on both protected and non-protected covariates; and  
(ii) a decision-making stage, in which a generalized distortion risk measure is applied to the target variable, conditional only on non-protected covariates, to determine the decision. In this second step we modify the risk measure such that the decision becomes insensitive to the protected attribute, thus enforcing fairness to ensure equitable outcomes under risk-sensitive, regulatory constraints. Furthermore, by utilising the concept of cascade sensitivity, we extend the marginal fairness framework to capture how dependencies between covariates propagate the influence of protected attributes through the modeling pipeline. A numerical study and an empirical implementation using an auto insurance dataset demonstrate how the framework can be applied in practice.

\end{abstract}

\begin{keywords}
Fairness, discrimination, distortion risk measures, sensitivity, insurance
\end{keywords}

\section{Introduction}
Ensuring fairness in algorithmic decision-making has become a central concern in high-stakes domains such as employment, credit scoring, and insurance. Traditional approaches to fairness predominantly focus on regulating machine learning predictions by controlling the use of protected attributes—such as gender, race, and nationality (e.g., \cite{dwork2012fairness,hardt2016equality,kusner2017counterfactual})—but often fall short when decision-making involves considerations beyond pure prediction. In insurance pricing for instance, recent works have primarily focused on imposing fairness constraints on pure premium models, which are a statistical or machine learning task \cite{Lindholm2022ASTIN, xin2024}. However, insurance prices typically incorporate not only expected losses (pure premium) but also risk margins and capital loadings, components governed by risk measures rather than pure expectations \cite{mildenhall2022pricing}. Thus these elements are not captured by fairness frameworks that rely solely on expected loss modeling.

Recently fairness notions have been extended beyond predictions by incorporating fairness considerations into business decision-making. For example, \cite{cohen2022price} and \cite{yang2024fairness} study fairness constraints in price discrimination primarily motivated by retail, e-commerce, and consumer service applications. These approaches rely on settings where firms optimize prices based on demand modeling—a practice that has been prohibited in some jurisdictions for highly regulated industries, such as insurance and finance \cite{POWhitePaper,shimao2022welfare}. Moreover, both \cite{cohen2022price} and \cite{yang2024fairness} implement fairness constraints in pricing decisions by conditioning on consumer valuations and demand functions, without modeling cost uncertainty or incorporating risk-based decision-making, making them less applicable to stochastic cost industries, such as insurance, where risk-adjusted decision-making and regulatory capital considerations are essential \cite{zhang2024fairness}. In these settings, decisions typically rely on risk measures, an area that remains to a large extent underexplored in the fairness literature. This motivates our first question:
\textit{How can we achieve fairness in decision-making with risk measures?}

Real-world pricing practice often involves two distinct stages: a modeling stage, where protected attributes may be used for (internal) risk assessment, followed by a decision-making stage, where decisions, e.g., insurance premiums, are determined and fairness regulations apply. In many highly regulated industries, direct discrimination using protected attributes at the decision-making stage is prohibited. A compelling example arises from the European Union’s gender-neutral pricing regulation (Directive 2004/113/EC (``Gender Directive'')): insurers are permitted to use gender when modeling claim costs, but it is prohibited that gender influence premiums, which include both expected losses and risk margins. The industry’s common response—simply removing gender from  decision-making —reflects the principle of ``fairness through unawareness.'' Yet this approach is widely recognized as ineffective, as the protected variable's influence may persist in the decision and result in indirect discrimination. This observation motivates our second question:
\textit{How can we ensure fairness in decision-making with risk measures, while allowing protected attributes in the modeling stage?}

To address these questions, we propose a \textit{marginal fairness} framework tailored to highly regulated and risk-sensitive sectors and conceptualize decision-making as a two-step process:  
(i) a \textit{predictive modeling stage}, where the prediction function for the target variable (e.g., insurance loss) is estimated using both protected (potentially discriminatory) attributes and permissible (non-protected) covariates; followed by  
(ii) a \textit{decision-making stage}, where a generalized distortion risk measure is applied to determine the risk-adjusted decision using only the permissible attributes. In this second step, we propose to modify the decision rule, such that the decision becomes insensitive to the protected attributes.
This approach allows protected attributes to inform accurate risk assessment while ensuring that they do not influence decisions. By embedding fairness at the decision layer and as decisions are based on risk measures, our framework extends fairness beyond mean-based predictions and profit-driven price optimization, offering a unified approach for fairness in settings governed by regulation and risk management. We summarize our contributions as follows:

\begin{enumerate}[label = $\roman*)$]
    \item \textit{A new individual fairness criterion—marginal fairness—for decision-making with risk measures:} We define marginal fairness as the insensitivity of decisions to small perturbations in protected attributes. By adopting a two-step decision-making process—prediction followed by risk-based decision—and introducing both marginal sensitivity and cascade sensitivity, this criterion offers new perspectives on mitigating indirect discrimination and comprehensively addresses both direct and indirect discrimination. This fairness notion aligns with regulatory standards such as the EU Gender-Neutral Pricing Directive, making it particularly relevant for applications in insurance and related domains.
    
 \item \textit{A consistent and unified analytical framework for achieving marginal fairness:} We provide a general theorem for deriving marginally fair decision rules across a wide range of practical settings, including cases where protected attributes are continuous, bounded, discrete, categorical, or multivariate. The core structure of the theorem remains unchanged; only the sensitivity measure needs to be adapted for each scenario, ensuring broad applicability in real-world decision problems.

    \item \textit{A practical implementation for empirical studies:} Using auto insurance data, we demonstrate how marginal fairness is applied in practice and compare against existing approaches, including fairness through unawareness and discrimination-free pricing \cite{Lindholm2022ASTIN,pope2011implementing}.
\end{enumerate}

\subsection{Related works} 

Fairness in machine learning has become a central research topic over the past decade, with a proliferation of fairness criteria and algorithmic interventions aimed at mitigating discrimination in automated systems. A comprehensive overview is provided in \cite{barocas-hardt-narayanan}. Broadly, fairness notions in the machine learning literature fall into two categories: group fairness and individual fairness. Group fairness, exemplified by demographic parity (or statistical parity), requires equality of outcomes across protected groups. Individual fairness, introduced by \cite{dwork2012fairness}, is based on the principle of “treating similar individuals similarly.” These two notions are often in tension, as demonstrated in \cite{binns2020apparent}, which explores the inherent trade-offs between group- and individual-level objectives. Our paper contributes to the individual fairness literature by introducing marginal fairness, that ensures decisions based on generalized distortion risk measures are insensitive to protected attributes. This aligns with regulatory frameworks such as the EU’s gender-neutral pricing directive and departs from prior approaches that enforce fairness at the prediction stage and focus on the expected values of a target variable.

A foundational approach to individual fairness in machine learning is proposed by \cite{dwork2012fairness}, who formalize the principle of “treating similar individuals similarly” through the notion of fairness through awareness. Their framework requires a predefined task-specific similarity metric over individuals, and fairness is enforced by ensuring that the decision function is Lipschitz continuous with respect to this metric. While elegant in theory, this approach relies on the availability of a suitable and ethically accepted similarity metric, which may be difficult to define in practice. Building on this idea, \cite{singh2023sensitive} propose a method for learning a sensitive subspace that captures variation associated with protected attributes. They enforce robustness of predictions to perturbations within this subspace, achieving a form of individual fairness without requiring an explicit similarity metric. However, neither of these approaches relies on derivative-based sensitivity analysis as we do: Dwork et al. use a global smoothness constraint based on pairwise distances, while Singh et al. use perturbation-based robustness in latent space. In contrast, our work enforces fairness by eliminating derivative-based sensitivity of the decision to protected attributes. This address both direct and indirect discrimination in situations when fairness must be ensured in the decision outcome rather than solely in the prediction.

Recent literature on pricing discrimination has predominantly focused on business sectors such as retail, e-commerce, and platform services, where firms optimize prices to maximize profits under fairness constraints. For example, \cite{cohen2022price} develop a profit-maximization framework with fairness-imposed adjustments to price discrimination strategies, aligning closely with applications in personalized retail pricing and online marketplaces. Similarly, \cite{yang2024fairness} extend this line of work to competitive markets, studying the impact of fairness regulations on pricing strategies in a duopoly setting. Both approaches retain profit maximization as the primary objective, treating fairness as an external constraint imposed on firm behavior. In contrast, our fairness framework is suited to industries where pricing decisions must follow regulatory and solvency requirements, and where price optimization may be restricted or prohibited. For example, in the United States, around 20 states have implemented price optimization bans since 2015, prohibiting insurers from using sophisticated data mining tools and modeling techniques during the rate-making process based on factors unrelated to a person's risk \cite{POWhitePaper}. In the United Kingdom, the Financial Conduct Authority (FCA) banned insurers from charging higher prices for renewals than for risk-identical new customers \cite{FCA2021Rule}. 

The literature on fair insurance pricing in the actuarial domain has largely focused on fairness in cost modeling—also known as pure premium pricing—where fairness constraints are imposed on statistical or machine learning models used to predict loss costs; see, e.g., \cite{Lindholm2022ASTIN,pope2011implementing,Frees2023,araiza2022discrimination, cote2024fair}. However, ensuring fairness in predicted loss costs alone does not guarantee fairness in pricing outcomes. In practice, insurance pricing incorporates additional components such as profit loadings and capital charges, which go beyond pure prediction. Recent research has therefore begun to explore fairness at the decision-making stage. For instance, \cite{shimao2022welfare} examine the welfare implications of fair pricing regulations through a comprehensive framework that includes cost modeling, demand modeling, and price optimization. Unlike \cite{shimao2022welfare}, we do not optimize over prices; instead, we assume that price optimization is restricted or prohibited, consistent with regulatory practices in, e.g., insurance markets. \cite{huang2025learning} introduce a two-step decision-making framework that separates predictive modeling—via factor models—from pricing decisions, and apply a decision error parity as the fairness criterion. While \cite{huang2025learning} allow for direct discrimination in both the modeling and decision stages, in our setting direct discrimination is absent at the decision stage. 

Compared to most literature in fair insurance pricing, which largely ignores risk margins, we employ generalized distortion risk measures to model decisions, thereby extending the notion of fair pure premium to fair technical premium that captures both expected loss and risk margin in a fairness-aware manner. \cite{zhang2024fairness} consider fairness in the loading component of catastrophe insurance pricing based on a specific set of axioms. However, the fairness axioms in their framework are not explicitly connected to protected attributes. In contrast, we consider a broad class of generalized distortion risk measures, including classical risk measures such as Expected Shortfall (also called Conditional Value-at-Risk), and define fairness via the elimination of sensitivity to protected attributes, thereby aligning more closely with industry practice \cite{mildenhall2022pricing} and regulatory fairness concerns.

This paper also connects to the growing literature on fairness risk measures, particularly the framework of \cite{williamson2019fairness}, where fairness is incorporated into the model training process via risk-sensitive loss functions. In that approach, risk measures such as Expected Shortfall are applied to the distribution of subgroup-specific losses to ensure robustness against poor outcomes for disadvantaged groups.  By contrast, our framework applies risk measures directly to the predicted outcomes, shaping the decision rule itself (e.g., a premium or loan price). This aligns with real-world practice in domains like insurance and finance, where decisions are often risk-adjusted.

We further link to the growing literature on sensitivity-based analysis, which offers tools to quantify how model outputs respond to changes in inputs. Foundational contributions such as \cite{Tsanakas2016RA} and \cite{Borgonovo2021EJOR} develop probabilistic and risk-based sensitivity measures to assess the value and influence of input variables. These techniques have been extended to more complex model structures, including discontinuities and discrete covariates \cite{Pesenti2024EJOR-non-diff}. While these approaches were not originally designed for fairness, they have recently inspired a range of fairness-aware methodologies.
Of particular relevance is the work of \cite{lindholm2024sensitivity}, who use variance-based sensitivity analysis to attribute proxy discrimination, originally formalized in the discrimination-free pricing framework of \cite{Lindholm2022ASTIN}, to individual covariates. Similarly, \cite{watson2022global} apply variance-based global sensitivity analysis to assess the overall influence of features on model outputs, linking these measures to notions of group fairness. In contrast, our proposed concept of marginal fairness targets the derivative-based sensitivity of the decision rule to protected attributes, rather than evaluating residual variance in predictive models. This distinction is especially important in applications where decisions are informed by generalized distortion risk measures, and fairness must be enforced at the outcome level, and not at the modeling stage.

\subsection{Road map}
The remainder of this paper is organized as follows. \Cref{sec:decision-framework} introduces the two-step framework for decision-making under generalized distortion risk measures, separating predictive modeling from risk-based decisions. \Cref{sec:MarginalFairness} formally defines marginal fairness as an individual fairness criterion appropriate for regulated decision-making environments. \Cref{sec:AchieveMF} develops theoretical results for achieving marginal fairness, including the characterization of optimal fair decision rules. The proposed framework accommodates various types of protected attributes, including continuous, bounded, discrete, categorical, and multivariate variables. \Cref{sec:cascade} introduces the concept of cascade sensitivity to mitigate indirect discrimination arising from statistical dependencies among covariates. \Cref{sec:simulation} presents a numerical study while \Cref{sec:empirical} describes the empirical implementation of marginally fair decisions using a French auto insurance dataset and benchmarks the results against alternative fairness strategies. \Cref{sec:conclusion} concludes the paper with a discussion of limitations and directions for future research. \Cref{app:ex} collects additional details on examples, \Cref{app:aux-results} states auxiliary results needed for proofs, which are all delegated go \Cref{app:proofs}. \Cref{app:model} collects further details on the empirical implementation.

\section{Decision-making with generalized distortion risk measures}\label{sec:decision-framework}

In many real-world applications,  such as insurance, finance, and public policy, decisions are not based solely on expected outcomes but also incorporate risk aversion, regulatory capital requirements, or profit loadings. Generalized distortion risk measures provide a flexible and interpretable framework for modeling such decisions.

\subsection{Risk informed decisions}
Our setup is a probability space $(\Omega, \F, \P)$ and we denote by $\L := L^2(\Omega, \F, \P)$ the space of square-integrable random variables (rvs). For a random vector $\Z := (Z_1, \ldots, Z_K)$, $K \in \N$, we denote its cumulative distribution function (cdf) by $F_\Z(\z) := \P(\Z \le \z)$, where $\z := (z_1, \ldots, z_K) \in \R^K$. For a univariate rv $Z$, we denote by $\Finv_Z(u) := \inf\{z \in \R \,|\, F_Z(z) \ge u\}$, $u \in (0,1)$, its (left-continuous) quantile function. We use the abbreviation $\P$-a.s. to mean $\P$-almost surely.

We consider an agent making a decision based on a univariate response variable $Y \in \L$, modeled through two types of covariates: $\D := (D_1, \ldots, D_m)$ (protected) and $\X := (X_1, \ldots, X_n)$ (non-protected), with $m,n \in \N$. We assume the split between $\D$ and $\X$ is exogenously determined (e.g., by legislation or regulation). The response is partially explained by these covariates via a prediction function $\mfg \colon \R^{m+n} \to \R$, such that
\[
Y := \mfg(\D, \X) + \ep\,,
\]
where $\ep$ is a noise term independent of $(\D, \X)$. Since $\ep$ is not relevant for the exposition, we simply write $Y := \mfg(\D, \X)$. A decision maker then applies a generalized distortion risk measure to inform decisions, recalled next.
\begin{definition}[Generalized distortion risk measure]
A generalized distortion risk measure $\rho_h \colon \L \to \R$ is defined as a signed Choquet integral
\begin{align*}
\rho_h(Y) := 
- \int_{-\infty}^0 \big( h(1) - h(1 - F_Y(x)) \big) \, \diff x 
+ \int_0^{\infty} h(1 - F_Y(x)) \, \diff x\,,
\end{align*}
where $h \in \mcH$ is a generalized distortion function from the class
\[
\mcH := \left\{ h \colon (0,1) \to \R ~\middle|~ h(0) = 0,\, \text{absolutely continuous},\, \int_0^1 (h'(u))^2 \, \diff u < \infty \right\}.
\]
\end{definition}

By Proposition 1 in \cite{Wang2020ASTIN}, generalized distortion risk measures are finite-valued, that is $\rho_h(Y)<+\infty$ for all $Y \in \L$ and all $h \in \mcH$. This class includes classical distortion risk measures where $h$ is non-decreasing, maps to $[0,1]$, and satisfies $h(1)=1$, encompassing well-known metrics such as Expected Shortfall (ES), power distortions, and inverse S-shaped distortions \cite{yaari1987Econometrica}. Furthermore, the family of distortion risk measures span the class of comonotonic additive and coherent risk measures \cite{Kusuoka2001AME}. Generalized distortion risk measures also include deviation measures such as the Gini deviation, inter-quantile range, and inter-ES range.

A generalized distortion risk measure $\rho_h(Y)$ has representation:
\begin{equation}\label{eqn:distortion-rm-rep}
        \rho_h(Y)
        =
        \int_0^1 \Finv_Y(u) \gamma(u)\, \diff u
        =
        \E[Y \gamma (F_Y(Y))]
        =
        \E[Y \gamma (U_Y)]
        \,,
    \end{equation} 
    where $U_Y:= F_Y(Y)\sim U (0,1)$ is a uniform rv that is comonotonic with $Y$, and $\gamma(u) := \partial_- h(x)|_{x = 1-u}$, $u \in (0,1)$, is the \emph{(generalized) weight function}, and $\partial_-$ denotes the left-derivative \cite{Wang2020ASTIN}.
    The weight function $\gamma(\cdot)$ provides insight into how risk is weighted across different quantiles. We assume throughout that the generalized distortion risk measures have representation  \eqref{eqn:distortion-rm-rep} and write with slight abuse of notation $\rho_\gamma$ instead of $\rho_h$.

\begin{example}[Expected Shortfall]
Expected Shortfall (ES) at level $\alpha \in [0,1)$ is a special case of a generalized distortion risk measure
\[
\ES_\alpha(Y) := \frac{1}{1 - \alpha} \int_\alpha^1 \Finv_Y(u) \, \diff u,
\]
with weight function $\gamma(u) = \frac{1}{1 - \alpha} \Id_{\{u \ge \alpha\}}$.
\end{example}

Any generalized distortion risk measure can be decomposed into an expected value (interpretable as a predictive value or best-estimate price in insurance), $\E[Y]$, and a risk loading, $\rho_{\tilde{\gamma}}(Y)$ defined below. For  $\gamma \in \Gamma$, define $\tilde{\gamma}(u) := \gamma(u) - 1$, $u \in (0,1)$, then the generalized distortion risk measure satisfies
\begin{equation}\label{eq:rm-decomp}
\rho_\gamma(Y) = \E[Y] + \rho_{\tilde{\gamma}}(Y) = \E[Y] + \int_0^1 \Finv_Y(u) \tilde{\gamma}(u)\, \diff u,
\end{equation}
Thus, making generalized distortion risk measures fair, implicitly implies that the sum of the expected value and the risk margin are fair.

\begin{example}[Discrimination in insurance]
Our setup aligns with insurance frameworks for discrimination-free models in \cite{Lindholm2022ASTIN,pope2011implementing}, where $Y$ represents policyholder claims and $\D$ includes protected characteristics such as gender or race. Most existing literature imposes fairness in the conditional expected value $\E[Y|\X]$, while we focus on fairness in the conditional \emph{risk-adjusted premium} $\rho_\gamma(Y|\X)$. By \eqref{eq:rm-decomp}, a generalized distortion risk measure decomposes into the best-estimate and a risk margin. This aligns with the natural allocation principle used in actuarial pricing, where the risk margin is distributed across policyholders to determine the technical premium \cite{tsanakas2009to}.
\end{example}

\subsection{Fairness in decision-making}
To incorporate fairness in decision-making, we formulate the decision-making process in two steps that align with real-world applications and regulatory requirements. First, we perform a predictive modeling task to estimate the prediction function for the target variable (e.g. insurance loss) $Y = \mfg(\D, \X)$, using both protected and non-protected variables. Second, we use a generalized distortion risk measure $\rho_\gamma(Y|\X)$ to determine the decision based only on the non-protected variables $\X$. Note that in general, fairness can be applied at either stage, giving rise to two distinct fairness notions:
\begin{definition}[Notions of fairness]
\begin{enumerate}[label = $\roman*)$]
    \item  \textit{Prediction fairness} arises when fairness constraints are applied to the predictive modeling step (e.g., ensuring the estimated target variable or prediction function $\hat{Y}=\hat{g}(\X,\D)$ is fair).
    \item \textit{Decision fairness} arises when fairness constraints are applied to the decision-making step (e.g., ensuring the estimated decision $\hat{\rho}_\gamma(Y|\X)$ is fair). 
\end{enumerate}
\end{definition}

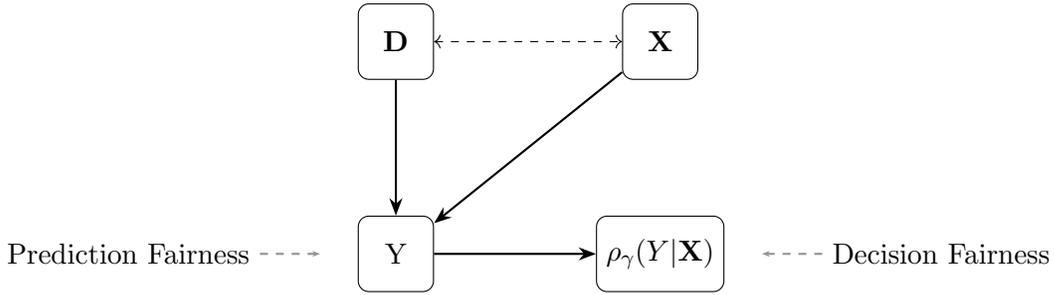
\begin{figure}[h]
\centering
\begin{tikzpicture}[
  node distance=1.8cm and 2.5cm,
  every node/.style={draw, minimum size=1cm, rounded corners},
  arrow/.style={-{Stealth}, thick},
  dashedarrow/.style={-{Stealth}, thick, dashed},
  annotation/.style={-{Stealth[length=3pt]}, thick, dashed, draw=gray},
  txt/.style={draw=none, fill=none}
]

\node (D) {$\D$};
\node[right=of D] (X) {$\X$};
\node[below=of D] (Y) {Y};
\node[below=of X] (rho) {$\rho_{\gamma}(Y|\X)$};

\node[txt, left=1.3cm of Y] (PFtxt) {Prediction Fairness};
\node[txt, right=1.3cm of rho] (DFtxt) {Decision Fairness};

\draw[<->, dashed] (D) -- (X);
\draw[arrow] (D) -- (Y);
\draw[arrow] (X) -- (Y);
\draw[arrow] (Y) -- (rho);
\draw[annotation] (PFtxt.east) -- ++(0.8, 0);
\draw[annotation] (DFtxt.west) -- ++(-0.8, 0);

\end{tikzpicture}

\caption{A graphical representation of the decision process (arrows indicate statistical or functional dependence, not causality). The predicted outcome $Y$ is modeled as a function of both protected attributes $\D$ and non-protected covariates $\X$. The decision \( \rho_\gamma(Y \mid \X) \) is a function of the conditional distribution of \( Y \) given \( \X \), which is modeled as a function of only $\X$. The dashed arrow between $\D$ and $\X$ indicates that dependence between $\D$ and $\X$ may or may not exist.}
\label{fig:decision-graph}
\end{figure}

\Cref{fig:decision-graph} illustrates the structure of our two-step decision framework. While the decision rule $\rho_\gamma(Y| \X)$ is a function only of the non-protected covariates (thus avoids direct discrimination), indirect discrimination may arise due to statistical dependence between $\D$ and $Y$, and potentially between $\D$ and $\X$. This highlights the need for fairness criteria that go beyond excluding protected variables. Prediction fairness addresses fairness at the modeling stage (i.e., ensuring $Y$ is not biased with respect to $\D$), while in our case decision fairness ensures that the decision is insensitive to distributional perturbations in protected attributes. The dashed arrow in \Cref{fig:decision-graph} from $\D$ to $\X$ indicates that statistical dependence between protected and non-protected covariates may or may not be present in a given application. Importantly, indirect discrimination can occur even if $\D$ and $\X$ are statistically independent, due to the dependence of $Y$ on $\D$.
We emphasize that this is not a causal graph, the arrows in \Cref{fig:decision-graph} represent statistical dependence or modeling structure, not causal relationships. Accordingly, the fairness criterion proposed in this paper does not rely on causal assumptions or counterfactual reasoning.

\textit{Direct discrimination} can be avoided by ensuring that protected attributes are not explicitly used in the decision-making process \cite{xin2024}. In our setting, this is satisfied as decisions are made on $Y$ conditional on non-protected covariates $\X$. In other words, individuals who differ only in protected attributes receive identical decision outcomes. A notable example of this principle is the European Union’s gender-neutral pricing regulation in insurance, which mandates that insurers must not use gender as a factor in determining individuals’ premiums and benefits in insurance contracts.

\textit{Indirect discrimination} has been defined in various ways across the literature \cite{Lindholm2022ASTIN,xin2024}. It is well understood that no single algorithm can simultaneously satisfy all proposed fairness criteria for mitigating (indirect) discrimination, except under strong and often unrealistic constraints \cite{hedden2021statistical,kleinberg2016inherent}. In this paper, we say indirect discrimination occurs, if a distributional perturbation of a protected attribute leads to a change in the decision.
In other words, indirect discrimination arises when protected attributes have value of information for decision-making, even if they are formally excluded from the decision rule. Due to the statistical dependence between the protected attributes \(\D\), the target variable \(Y\), and the non-protected covariates \(\X\), decisions can be affected by \(\D\) through indirect pathways. The marginal fairness framework addresses this in two ways. First, by eliminating the sensitivity to protected attributes, while holding the data-generating process fixed, (that is path $\D\to Y\to \rho_\gamma(Y|\X)$ in \Cref{fig:decision-graph}) and second by allowing perturbations of the protected attributes to impact non-protected attributes, thus additionally accounting for the indirect path $\D\to \X\to Y\to \rho_\gamma(Y|\X)$ in \Cref{fig:decision-graph}. Thus, by eliminating the sensitivity of decision outcomes to protected attributes, the no value of information condition is enforced: the distribution of decisions remains stable under small perturbations to the distribution of protected attributes. We refer to \cite{Borgonovo2021EJOR} and \cite{Fissler2023EJOR} for related discussions on the connections between sensitivity analysis and value of information concepts in model evaluation. Fairness, in this view, is achieved not simply by excluding \(\D\) from the decision rule, but by ensuring its functional irrelevance.

\section{Marginal fairness} \label{sec:MarginalFairness}

The agent considers a decision rule given by a generalized distortion risk measure applied to the target $Y$ conditional on the non-protected covariates $\X$. For this we denote by $F_{Y|\X}(\cdot\,|\,\x):= \P(Y \le \cdot\,|\,\X = \x)$ and by $\Finv_{Y|\X}(u\,|\,\x):= \inf\{y \in \R\,|\, F_{Y|\X}(y\,|\,\x) \le u\}$, $u \in (0,1)$, the conditional cdf and conditional quantile function of $Y$ given $\X = \x$, respectively. Then by \eqref{eqn:distortion-rm-rep} it holds a.s. that
\begin{equation}\label{eq:decision-rule}
\rho_\gamma(\,Y\,|\,\X\,) 
=
\int_0^1 \Finv_{Y|\X}(u\,|\,\X) \gamma(u) \diff u
=
\E\big[Y \gamma \big(F_{Y|\X}(Y\,|\,\X)\big)\, |\X\big]
=
\E\big[Y \gamma (U_{Y|\X})\, |\X\big]
\,,
\end{equation} 
where $U_{Y|\X}:= F_{Y|\X}(Y\,|\,\X)$ is a uniform random variable on $(0,1)$.

Even when protected attributes are excluded from the decision rule $\rho_\gamma(Y|\X)$ (no direct discrimination), statistical dependence between $\D$ and $Y$, and potentially between $\D$ and $\X$ may still lead to unfair outcomes in $\rho_\gamma(Y|\X)$ (indirect discrimination). To capture these effects, we propose a fairness criterion based on \emph{sensitivity} of the decision rule to protected attributes, which we term \emph{marginal fairness}. The idea of marginal fairness is that small perturbations in protected attributes do not affect the decision. 

\begin{definition}[Marginal fairness]
A decision rule $\rho_\gamma$ is marginally fair for covariate $D_i$ if
\begin{equation*}
    \partial_{D_i}\;  \rho_\gamma(\, Y\,|\,\X\,) = 0\,, \quad \P\text{-a.s.,}
\end{equation*}
where the partial derivative is defined by
\begin{equation*}
    \partial_{D_i}\;  \rho_\gamma(\, Y\,|\,\X\,):= \lim_{\delta \to 0}\frac{
    \rho_\gamma\big( \mfg\big( \D_{i,\delta} , \X\big)\,|\,\X\big) - \rho_\gamma\big( \mfg\big( \D,\X\big)\,|\,\X\big)}{\delta}\,,
\end{equation*}
with $\D_{i,\delta}:= (D_1, \ldots, D_{i-1}, D_{i,\delta}, D_{i+1}, \ldots, D_m)$ and for a perturbation $D_{i,\delta}$.
\end{definition}

The term $\partial_{D_i}\;  \rho_\gamma(\, Y\,|\,\X\,)$ is the differential sensitivity (G\^ateaux derivative) of $\rho_\gamma(\cdot)$ applied to the conditional rv $Y|\X$, in direction of the protected attribute $D_i$; indicatively see \cite{Tsanakas2016RA}. 
A marginally fair decision rule exhibits zero sensitivity to the protected covariate \(D_i\), meaning that the protected attribute has no value of information for decision-making: small perturbations in the distribution of \(D_i\) do not affect the decision rule. While a variety of sensitivity analysis techniques exist—including Sobol indices, score-based sensitivity for elicitable functionals \cite{Fissler2023EJOR}, moment-independent measures \cite{Borgonovo2016RA}, quantile-based sensitivity \cite{Tsanakas2016RA}, and indices based on optimal transport \cite{borgonovo2024MS}—we argue that the choice of sensitivity measure should reflect the nature of the decision statistic. In our context, that statistic is the generalized distortion risk measure applied to $Y|\X$. We therefore adopt a differential sensitivity approach, which measures the response of the decision rule to infinitesimal perturbations in $D_i$. This framework is applicable across the entire class of generalized distortion risk measures and provides a natural link between fairness and the robustness of decisions. It ensures that the decision rule remains stable under minor shifts in the distribution of protected variables.

The choice of perturbation for a protected covariate $D_i$ should depend on its support, denoted by $\supp(D_i)$. For instance, if $D_i$ is a continuous variable with support on $\R$, a natural choice is a \emph{proportional perturbation} of the form $D_{i,\delta} = D_i(1 + \delta)$ for small $\delta > 0$. This perturbs all values of $D_i$ multiplicatively, which can be interpreted as distorting the scale of the distribution—for example, modifying its standard deviation. From a distributional perspective, this perturbation transforms the cumulative distribution function (cdf) of $D_i$ to $F_{D_i}(\cdot / (1 + \delta))$. This viewpoint interprets the perturbation as a cdf distortion, and we refer to \cite{Pesenti2024EJOR-non-diff} for an extensive treatment of such transformations.
If $D_i$ is instead a categorical or discrete random variable, proportional perturbation is no longer appropriate, and alternative approaches must be used to preserve the support of $D_i$. To simplify the exposition, we first discuss protected covariates $D_i$, $i = 1, \ldots, m$, that are supported on $\R$. We then address the case of bounded support in \Cref{sec:bouded-support}, followed by the treatment of discrete covariates in \Cref{sec:discrete}.

In general a decision rule \eqref{eq:decision-rule} is not marginally fair with respect to covariate $D_i$ as can be seen in the following illustrative example.

\begin{example}\label{ex:mean-linear}
For simplicity, consider an insurance setting with two covariates as rating factors, $X$ and $D$, i.e., $n = m = 1$, and a linear model for the claims cost
\[
Y = \beta_0 + \beta_1 X + \beta_2 D + \ep \quad \text{a.s.,}
\]
where $\beta_0, \beta_1, \beta_2 \in \R \setminus \{0\}$, and $\ep \sim \mathcal{N}(0,1)$ is independent of $(D, X)$. We take the expected value as the decision rule for pure premium estimation, i.e., $\rho_1$ with weight function $\gamma(u) \equiv 1$, in which case
\begin{equation}\label{ex:eq:unaware-price}
    \E[Y \mid X = x] = \beta_0 + \beta_1 x + \beta_2\, \E[D \mid X = x].
\end{equation}
Consider a proportional perturbation $D_\delta := D (1 + \delta)$, which yields a perturbed outcome 
\[
Y_\delta := \beta_0 + \beta_1 X + \beta_2 D_\delta + \ep.
\]
The sensitivity to $D$ can then be computed as
\begin{align*}
    \partial_{D}\, \E[Y \mid X = x] 
    &= \lim_{\delta \to 0} \frac{1}{\delta} \left( \E[Y_\delta \mid X = x] - \E[Y \mid X = x] \right) \\
    &= \beta_2 \lim_{\delta \to 0} \frac{1}{\delta} \left( \E[D(1 + \delta) \mid X = x] - \E[D \mid X = x] \right) \\
    &= \beta_2\, \E[D \mid X = x].
\end{align*}

Hence, the conditional expectation is not marginally fair unless $\E[D \mid X = x] = 0$ for all $x \in \R$. In the insurance literature, the pricing rule \eqref{ex:eq:unaware-price} is often referred to as the \emph{unawareness price}, as it ignores the explicit use of $D$ in the decision rule. However, it is well-established that this approach can result in discrimination with respect to $D$; see, for instance, \cite{Lindholm2022ASTIN}.
Note that even if $(X, D)$ are statistically independent, the sensitivity to $D$ does not necessarily vanish, as $Y$ is still directly influenced by $D$, as shown in \Cref{fig:decision-graph}.
\end{example}

In practice, there may be multiple protected attributes, such as gender, race, and religion. Marginal fairness can be generalized to include fairness with respect to all protected covariates, a notion we term \emph{multi-marginal fairness}.

\begin{definition}[Multi-marginal fairness]
    A decision rule is multi-marginally fair if it is marginally fair for all covariates $D_i$, $i = 1\ldots, m$.
\end{definition}

To establish marginal fairness, we need a succinct representation of the sensitivity of the decision rule with respect to a protected covariate, which is established below.

\begin{proposition}[Marginal sensitivity]\label{prop:marginal-sensitivity}
Let $\supp(D_i) = \R$ and consider the perturbation $D_{i,\delta} = D_i(1 + \delta)$. Assume that $\mfg$ is invertible in the $i$-th component and that for all $u \in (0,1)$, the function $\delta \to \Finv_{\mfg\big( \D_{i,\delta}, \X \big)}(u)$, $\delta \ge 0$, is differentiable in a neighborhood of $\delta = 0$ with bounded derivative.
    Then, it holds $\P$-a.s. that 
    \begin{align*}
        \partial_{D_i}\;  \rho_\gamma(\, Y\,|\,\X\,)
        &=
        \E\big[\,D_i \,\partial_{i} \mfg(\D, \X) \gamma\big(U_{Y|\X}\big)  \,|\, \X\,\big]\,,
    \end{align*}
where we define $\partial_k \mfg(z_1, \ldots, z_{m+n}):= \frac{\partial}{\partial z_k}\mfg(z_1, \ldots, z_{m+n})$ as the partial derivative of $\mfg$ with respect to its $k$-th component.
\end{proposition}

\Cref{prop:marginal-sensitivity} characterizes the marginal effect of a perturbation in a protected variable \( D_i \) on the decision rule \( \rho_\gamma(Y|\X) \). Intuitively, \( D_i \) influences the prediction function, which in turn affects how the outcome distribution is distorted in the decision rule. The derivative \( \partial_i \mfg(\D, \X) \) captures how this influence shifts the relative emphasis placed on different outcomes. The function \( \gamma \) reflects the decision-maker’s risk preferences, placing more weight on outcomes that are considered more critical (e.g., large losses). Together, these elements quantify how sensitive the decision is to variations in the protected attribute.

The next example illustrates that even if $Y$ and $D$ are dependent, the sensitivity of a distortion risk measure to $D$ can vanish. This happens if $D$ is irrelevant to the decision criterion.
\begin{example}\label{ex:Y-D-dependent-zero-sens}
Consider $Y = \Id_{\{X_1 = 0\}}D + \Id_{\{X_1 = 1\}} X_2$, where $X_1 \sim Ber(p)$, i.e. $\P(X_1 = 1) = 1 - \P(X_1 = 0) = p$, $D \in [0, C]$, for some $C \ge 0$, $X_2 > C$ $\P$-a.s., and $D, X_1, X_2$ are independent. Here, the protected covariate $D$ only affects $Y$ up to its quantile level of $1-p$; we also refer to \Cref{app:ex} for additional details on this example, e.g. the cdf and quantile function of $Y$.

If the decision rule is the unconditional ES at level $\alpha > 1-p$ --- recall that $D$ does not affect $\ES_\alpha(Y)$ --- the decision rule is marginally fair. Indeed the sensitivity to protected covariate $D$, applying \Cref{prop:marginal-sensitivity}, is
\begin{align*}
    \partial_{D}\;  \ES_\alpha(\, Y)
        &=
       \frac{1}{1-\alpha}\, \E\big[ \,\Id_{\{X_1 = 0\}} \,D\,\Id_{\{U_{Y} \ge \alpha\}}\big]\,.
\end{align*}
We obtain that $U_{Y} = F_Y(Y) \ge \alpha$ is equivalent to 
\begin{equation}\label{eq:ineq-indicator}
    F_Y(Y) = \Id_{\{Y \le C\}}\; (1-p)F_D(Y) + \Id_{\{Y >C\}}\; (1-p + p F_{X_2}(Y)) \ge \alpha.
\end{equation}
Moreover since $\P$-a.s. it holds that $(1-p)F_D(Y) \le (1-p) < \alpha$, \eqref{eq:ineq-indicator} is equivalent to $\Id_{\{X_1 = 1\}} \Id_{\{ F_{X_2}(Y) \ge \frac{\alpha - 1 + p}{p}\}}$\,.
Collecting, the sensitivity to $D$ is
\begin{align*}
    \partial_{D}\;  \ES_\alpha(\, Y)
        &=
       \frac{1}{1-\alpha}\, \E\big[ \,\Id_{\{X_1 = 0\}} \,D\,\Id_{\{X_1 = 1\}} \Id_{\{ F_{X_2}(Y) \ge \frac{\alpha - 1 + p}{p}\}}\big] = 0\,.
\end{align*}
If the decision rule is however the expected value or $\ES_\alpha$, with $\alpha \le 1-p$, then the decision rule is not marginally fair.

\end{example}

Our notion of marginal fairness relies on evaluating partial derivatives with respect to perturbations of a single protected covariate $D_i$, while holding all other covariates (and their joint dependence structure, i.e., their copula) fixed. This means that even if $D_i$ (e.g., nationality) is strongly correlated with another covariate $X_j$ (e.g., postal code), we consider only perturbations in $D_i$ while treating $X_j$ as fixed, thus mitigating indirect discrimination through the path $D \rightarrow Y \rightarrow \rho_\gamma(Y|\X)$ in \Cref{fig:decision-graph}. In \Cref{sec:cascade}, we generalize this concept to \emph{marginal fairness with cascade sensitivity}, where a perturbation in $D_i$ induces changes in $X_j$ and other covariates, according to their statistical dependence. This extension accounts for the full indirect influence of protected attributes on the decision rule through correlated features, that is also path $D \rightarrow \X \rightarrow Y \rightarrow \rho_\gamma(Y|\X)$ in \Cref{fig:decision-graph}.

\begin{remark}An important feature of our marginal fairness framework is that partial derivatives capture only \emph{infinitesimal perturbations} of the protected attribute $D_i$. These perturbations are interpreted as small distributional shifts in $D_i$, while preserving the dependence structure between $D_i$ and the remaining covariates $\X$—as characterized by their joint copula. As such, implausible or logically inconsistent combinations of $(\X, D_{i,\delta})$ do not arise. This ensures that the fairness analysis remains consistent with the observed data distribution and avoids hypothetical scenarios that may lack empirical support.\end{remark}

\subsection{Comparison with existing fairness criteria}
Marginal fairness differs fundamentally from many established fairness notions:
\begin{itemize}
\item \textit{Fairness through awareness\cite{dwork2012fairness}:} Marginal fairness is closely related to the notion of individual fairness—``treat similar individuals similarly''—as introduced by \cite{dwork2012fairness}. However, instead of relying on a pre-specified similarity metric between individuals, marginal fairness enforces a data-driven notion of fairness based on the decision rule's sensitivity to small, controlled perturbations in the distribution of protected attributes. It captures a form of fairness: if two individuals are similar in all non-protected attributes, then their decisions should not differ due to slight distributional changes in protected characteristics.

\item \textit{Demographic parity:} A group fairness criterion that requires statistical independence between the decision and protected attributes. In contrast, marginal fairness is an individual fairness criterion that allows for statistical dependence, but eliminates their sensitivity.


\item \textit{Counterfactual fairness 
\cite{kusner2017counterfactual}:} Defines fairness by requiring that a decision would remain unchanged in a counterfactual world where the protected attribute had been different, given a structural causal model. While both counterfactual fairness and marginal fairness address individual-level fairness, marginal fairness does not rely on causal assumptions. Instead, it evaluates fairness through the lens of distributional sensitivity—in a way that small changes in the distribution do not lead to implausible data combinations.

\item \textit{Variance-based sensitivity fairness \cite{lindholm2024sensitivity, watson2022global}:} Measures the global or residual influence of protected attributes using variance decomposition, focusing on expected values as the decision rule. In contrast, marginal fairness applies to the class of generalized distortion risk measures and is based on differential sensitivity.

\item \textit{Discrimination-free pricing \cite{Lindholm2022ASTIN,pope2011implementing}:} Mitigates proxy discrimination by averaging over the distribution of protected attributes in the predictive model, thereby avoiding omitted variable bias. In contrast, marginal fairness does not modify the predictive model directly, but instead modifies the decision rule, while keeping the underlying data generation fixed.
\end{itemize}

\section{Achieving marginal fairness} \label{sec:AchieveMF}

As seen in \Cref{ex:mean-linear}, decision rules derived from a generalized distortion risk measure \( \rho_\gamma \) are in general not marginally fair with respect to protected attributes. To address this, the decision maker seeks to construct a new risk measure \( \rho_\ell \) by minimally adjusting the weight function \( \gamma \), such that the resulting decision rule is marginally fair.

Mathematically, the decision maker solves
\begin{align}\label{opt:main}
\tag{P}
    \argmin_{\ell \in \Gamma^{\t, \x}} \int_0^1 \big(\gamma(u) - \ell(u)\big)^2\, \diff u
    \qquad \text{such that}\qquad 
    \partial_{D_i}\;  \rho_\ell(\, Y\,|\,\X\,)= 0\quad \P\text{-a.s.}\,,
\end{align}
where \( \Gamma^{\t, \x}\) denotes the class of square-integrable weight functions parametrized by $(\t,\x)$
\begin{equation*}
    \Gamma^{\t, \x}:= \Big\{ \ell^{\t, \x} \colon [0,1] \to \R \,\Big|\, \int_0^1 \big(\ell^{\t, \x}(u)\big)^2 \diff u <\infty  \Big\}\,.
\end{equation*}

The solution to \eqref{opt:main}, denoted \( \gamma^* \), defines a new generalized distortion risk measure \( \rho_{\gamma^*} \) that is marginally fair with respect to \( D_i \). Note that minimizing the squared \( L^2 \)-distance is a natural choice as generalized distortion risk measures require square-integrable weights, thus the objective introduces no additional restrictions while preserving interpretability and tractability. This approach achieves fairness by modifying the decision rule directly, rather than altering the distribution of the input covariates, which is in contrast to several works that impose fairness by distorting the joint distribution of the inputs \( (\D, \X) \) (see, e.g., \cite{Lindholm2022ASTIN}).

To establish the marginally fair decision rule we require mild integrability assumption, e.g., on the slope of the prediction function and the sensitivity. 
\begin{assumption}[Integrability]\label{asm-integrability}
Let $i \in \{1, \ldots, m\}$. The prediction function $\mfg$ is invertible in the $i$-th component and there exists constants $0<c_1, c_2$, such that for all $(\t, \x)\in \supp(\D,\X)$
\begin{align*}
    \big(t_i\, \partial_i \mfg(\t, \x)\big)^2 &< c_2 \,, 
    \\
    \big(\partial_{D_i}\;  \rho_\gamma(\, Y\,|\,\X = \x\,)\big)^2 &< c_2 \,,\quad \text{and}
    \\
    \E \big[\big( D_i \partial_i \mfg(\D, \X)\big)^2 \,|\, \X = \x \big] &> c_1  \,.
\end{align*}
\end{assumption}

\subsection{Continuous protected variables}\label{sec:continuous-rv}

In this section, we assume that the protected covariates have support \( \R \) and are continuously distributed. The following result characterizes marginally fair decision rules in this setting. While we focus here on continuously distributed covariates, we emphasize that the results also hold when \( D_i \) has compact support or is discrete; see \Cref{sec:bouded-support,sec:discrete}. The only difference lies in the specific expression of the differential sensitivity.

\begin{theorem}[Marginally fair decision rule]\label{thm:individual}
Let $i \in \{1, \ldots, m\}$ and let \Cref{asm-integrability} be satisfied for $i$. Then there exists a unique solution to optimization problem \eqref{opt:main} given by $\gamma^* \in \Gamma$ that satisfies 
   \begin{equation}
   \label{eq:optimal-h}
     \gamma^*(U_{Y|\X})=
    \gamma(U_{Y|\X}) - \frac{ \partial_{D_i}\;  \rho_\gamma(\, Y\,|\,\X\,)}{\E\big[\,\big(D_i \partial_{i} \mfg(\D, \X) \big)^2 |\,\X\big]}\;  D_i \, \partial_{i} \mfg(\D, \X)\,, \quad \P\text{-a.s.} \,.
   \end{equation}
Moreover, the unique marginally fair decision rule for covariate $D_i$ is
\begin{equation}\label{eq:marginally-fair}
    \rho^{D_i}_{\gamma^*}(Y \,|\,\X) 
    =
    \rho_\gamma(Y \,|\,\X) -
    \frac{ \partial_{D_i}\;  \rho_\gamma(\, Y\,|\,\X\,)}{\E\big[\,\big(D_i \partial_{i} \mfg(\D, \X) \big)^2 \big|\,\X\big]}\; 
    \E[Y \, D_i \, \partial_{i} \mfg(\D, \X)\, |\,\X ]
    \,.
\end{equation}
\end{theorem}

The expression in~\eqref{eq:marginally-fair} provides an explicit formula for the marginally fair decision rule associated with the adjusted weight function \( \gamma^* \). The first term, \( \rho_\gamma(Y \mid \X) \), represents the original (potentially unfair) decision rule defined by the generalized distortion risk measure. The second term is a correction that removes the influence of the protected covariate \( D_i \) on the decision. The numerator, \( \partial_{D_i} \rho_\gamma(Y \mid \X) \), quantifies how sensitive the original decision rule is to small perturbations in \( D_i \). The denominator, \( \E[(D_i  \partial_i \mfg(\D, \X))^2 \mid \X] \), acts as a normalization factor that captures the conditional variability of the influence of \( D_i \) on the prediction function \( \mfg(\D, \X) \). The expectation \( \E[Y D_i \, \partial_i \mfg(\D, \X) \mid \X] \) quantifies how the influence of the protected attribute \( D_i \) on the prediction function \( \mfg(\D, \X) \) interacts with the outcome \( Y \), conditional on the features \( \X \). It captures the extent to which variations in \( D_i \) not only affect the prediction but also co-vary with the outcome, thereby informing the appropriate direction and magnitude of the fairness correction.
Together, these terms yield a decision rule that is marginally fair—insensitive to changes in the protected variable—while remaining as close as possible to the original rule in terms of squared \( L^2 \)-distance.

Next, we discuss how to adjust the conditional expectation to achieve marginal fairness.

\begin{example}\label{ex:mean-linear-fair}
We continue \Cref{ex:mean-linear} and the marginally fair expected value becomes
\begin{align}\label{eq:mean-EV-linear}
    \rho^{D}_{1^*}(Y \,|\,X = x) 
        &=
    \tilde{\beta}_0(x) + \tilde{\beta}_1(x)\, x\,,
\end{align}
where $\tilde{\beta}_0(x) := \beta_0 \,(1 - c_x)$, $\tilde{\beta}_1(x) :=\beta_1  (1 - c_x)$, and $c_x:= \frac{\E[\, D | X= x]^2}{\E[\,D^2  |X= x]}$. Thus, the  constant coefficients $\beta_0, \beta_1, \beta_2$ are modified to be either functions of $x$ or vanish. 

This insight can be generalized to any generalized distortion risk measure $\rho_\gamma$. Indeed similar calculations show that the marginally fair decision rule for $\rho_\gamma$ is
\begin{equation}\label{eq:fair-decision-regression}
    \rho^{D}_{\gamma^*}(Y \,|\,X = x) 
        =
    \bar{\beta}_0(x)+ \bar{\beta}_1(x)\, x\,,
\end{equation}
where $\bar{\beta}_0(x) := \beta_0 \,(1 - \bar{c}_x) +  \beta_2\, \rho_{\gamma}(Y \,|\,X = x)- \partial_D \rho_{\gamma}(Y \,|\,X = x)  + \rho_\gamma(\ep)$, $\bar{\beta}_1(x) :=\beta_1  (1 - \bar{c}_x)$, and $\bar{c}_x:= \frac{ \partial_D \rho_{\gamma}(Y \,|\,X = x)\E[\, D | X= x]}{\beta_2\, \E[\,D^2  |X= x]}$.

In our framework, all coefficients $\beta_k$, $k = 0,1,2$, are distorted, which is in contrast to the discrimination-free price proposed in \cite{Lindholm2022ASTIN}, that changes the value multiplied with the coefficient of the protected variable, leaving $\beta_1$ fixed. 
\end{example}

A result similar to \Cref{prop:marginal-sensitivity} holds for multi-marginal fairness, however, the representation of the fair decision rule is only semi-explicit.

\begin{proposition}[Multi-marginally fair decision rules]
\label{prop:mutli-marginal}
Let \Cref{asm-integrability} be satisfied for all $i \in \{1, \ldots, m\}$. Then, if a multi-marginal fair decision rule exists, it is unique and given by 
    \begin{equation}\label{eq:muli-marginal-cont}
         \rho^{\D}_{\gamma^*}(Y \,|\,\X = \x) 
    =
    \rho_\gamma(Y \,|\,\X=\x) -
    \sum_{l = 1}^m \eta_l(\x)\;  
    \E[Y \, D_l \, \partial_{l} \mfg(\D, \X)\, |\,\X  = \x]\,,
    \end{equation}
where for each  $l = 1\ldots, m$ and each $\x \in \textup{\supp}(\X)$, the Lagrange parameters $\eta_l(\x) \in\R $ are such that
\begin{equation*}
     \partial_{D_l}\;  \rho^{\D}_{\gamma^*}(\, Y\,|\,\X\,)
     = 0\,,
     \quad \text{for all} \quad l \in \{1, \ldots, m\}\,.
\end{equation*}
\end{proposition}

\begin{remark}
    In some applications, the decision rule is unconditional, that is $\rho_\gamma(Y)$ rather than $\rho_\gamma(Y|\X = \x)$. Our framework extends to unconditional decision by defining marginal fairness via $\partial_{D_i}\, \rho_\gamma(Y) = 0$. Then, all results including propositions, theorems, and corollaries, apply when the conditioning on $\X$ is removed in the statements.
\end{remark}

\subsection{Continuous protected variables with bounded support}\label{sec:bouded-support}
In this section, we consider marginal fairness when the protected variable \( D_i \) has bounded support. Bounded random variables arise in many practical applications such as credit scores, age, or variables that lie within a fixed range. A key observation is that a perturbation of the form \( D_i(1 + \delta) \) may yield values outside the valid range of the covariate, making them unrealistic or operationally infeasible. To address this, we propose a cohesive perturbation that respects the bounded nature of such variables.

Recall that for a rv $U \sim U(0,1)$ comonotonic to $D_i$, it 
holds $\P$-a.s.
\begin{equation}\label{eq:Di-in-terms-of-U-compact}
    D_{i}= F_{D_i}^{-1} (U)
    =
    F_{D_i}^{-1} \left(\Phi \big(\Phi^{-1}(U)\big) \right)\,,
\end{equation}
 where $\Phi$, $\Phi^{-1}$ denote the standard normal cdf and quantile function, respectively. Similar to earlier sections, we consider a proportional perturbation, this time not directly on $D_i$ but on the standard normal rv $\Phi^{-1}(U)$ that generates $D_i$, i.e.,
\begin{equation}\label{eq:perturb-compact}
    D_{i,\delta}:= F_{D_i}^{-1} \left(\Phi \big(\Phi^{-1}(U) (1  + \delta)\big) \right)\,, \qquad \delta \ge 0\,.
\end{equation}
As $\Phi\big(\Phi^{-1}(u)(1+ \delta)\big) \in (0,1)$ for all $u \in (0,1)$, the perturbed rv has the same values as $D_i$ but distorted probabilities. Clearly for $\delta = 0$, we recover $D_{i,0} = D_i$, and moreover $\lim_{\delta \to 0}D_{i,\delta} = D_i$ holds $\P$-a.s.. Since $(D_{i}, D_{i, \delta})$ are comonotonic, the dependence (copula) of $(\D, \X)$ is equivalent to that of $(\D_{i,\delta}, \X)$.

Since the theorem on marginally fair decision rules relies on the sensitivity, we first derive the sensitivity to covariates that are compactly supported.

\begin{proposition}[Sensitivity - compact support]\label{prop:marginal-compact}
Let $D_i$ be an absolutely continuous random variable with compact support and perturbation given in \eqref{eq:perturb-compact}. Assume that $\mfg$ is invertible in the $i$-th component and that for all $u \in (0,1)$, the function $\delta \to \Finv_{\mfg\big( \D_{i,\delta}, \X \big)}(u)$ is differentiable in a neighborhood of $\delta = 0$ with bounded derivative.
    Then, it holds $\P$-a.s. that 
    \begin{equation*}
        \partial_{D_i}\;  \rho_\gamma(\, Y\,|\,\X\,)
        =
        \E\Big[\,\frac{\phi\big(\Phi^{-1}(F_{D_i}(D_i))\big)}{f_{d_i}(D_i)} \,\partial_{i} \mfg(\D, \X) \gamma\big(U_{Y|\X}\big)  \,|\, \X\,\Big]
        \,,
    \end{equation*}
    where $f_{d_i}(\cdot)$ is the density of $D_i$ and $\phi(\cdot)$ the standard normal density.
\end{proposition}

\Cref{prop:marginal-compact} characterizes how a perturbation of a continuous sensitive attribute \( D_i \) with compact support affects a distortion-based decision rule \( \rho_\gamma(Y|\X) \). The result shows that the sensitivity depends on three key components: the local effect of \( D_i \) on the model output (captured by \( \partial_i \mfg \)), the statistical weight of the individual's position within the distribution of \( D_i \) (through the ratio involving the standard normal density and the marginal density \( f_{d_i} \)), and the importance assigned to the individual’s outcome rank \( U_{Y|\X} \) via the distortion weight \( \gamma \). Intuitively, this means the influence of \( D_i \) on the decision is strongest where the model is locally sensitive, where the density of \( D_i \) is low (amplifying perturbations), and where the individual's outcome is given higher priority under the distortion function. 

With the sensitivity result for compactly supported covariates at hand, marginal fairness decision rules can be characterized in the same form as in the case of sensitive attributes with unbounded support, resulting in a unified framework.

\begin{corollary}[Marginal fairness - compact support]\label{cor:marginal-compact}
 Let $D_i$ be an absolutely continuous rv with compact support, the perturbation given in \eqref{eq:perturb-compact}, and assume that for all $u \in (0,1)$, the function $\delta \to \Finv_{\mfg\big( \D_{i,\delta}, \X \big)}(u)$ is differentiable in a neighborhood of $\delta = 0$ with bounded derivative. Then the following holds
\begin{enumerate}[label = $\roman*)$]
        \item the marginal fair decision rule is given in \Cref{thm:individual}\,, and
        \item the multi-marginal fair decision rule is given in \Cref{prop:mutli-marginal}, where each summand $l\in\{1, \ldots, m\}$ in \eqref{eq:muli-marginal-cont} is replaced by $$\eta_l(\x) \, \E\Big[\,Y \, \frac{\phi\big(\Phi^{-1}(F_{D_l}(D_l))\big)}{f_{d_l}(D_l)} \,\partial_{l} \mfg(\D, \X) \,|\, \X\,\Big]\,,$$
    \end{enumerate}
    and where the sensitivity to $D_i$ is given in \Cref{prop:marginal-compact}.
    \end{corollary}

\subsection{Discrete and categorical protected variables}\label{sec:discrete}

In this section, we generalize the marginal fairness framework to accommodate discrete and categorical protected variables. Many socially salient attributes—such as gender, race, and age—are either inherently discrete or can be discretized for analysis. Gender is often encoded as binary (e.g., male/female), race as a set of mutually exclusive categories (e.g., White, Black, Asian, etc.), and age as discrete groups or brackets (e.g., 18–25, 26–35, etc.). Moreover, categorical variables can be transformed into multiple discrete representations through methods such as one-hot encoding or embedding.

As we assume that $D_i$ is discrete, the perturbation \eqref{eq:Di-in-terms-of-U-compact} implies that the mapping $u \mapsto F_{D_i}^{-1} \left(\Phi \big(\Phi^{-1}(u)\big) \right)$ is discontinuous, making \Cref{prop:marginal-compact} inapplicable. Therefore, we apply techniques developed in Section 4 of \cite{Pesenti2024EJOR-non-diff} to derive the sensitivity to discrete random variables.

We recall the generalized distributional transform, which represents a discrete random variable in terms of a uniform random variable. Specifically, it holds $\P$-a.s. that
\begin{equation}\label{eq:Di-in-terms-of-U-discrete}
    D_{i}= F_{D_i}^{-1} (\tilde{U})\,,
\end{equation}
 where $\tilde{U}:= \tilde{F}_{D_i}\big(D_i; V\big)$, with $V \sim U(0,1)$ independent of $D_i$, is uniformly distributed on $(0,1)$. Moreover $\tilde{F}_{D_i}(t; \lambda):= \P(D_i< t) + \lambda\, \P(D_i = t)$, $\lambda \in (0,1)$, is the generalized distributional transform of $D_i$, see e.g., \cite{ruschendorf2013book}. Intuitively, the generalized distributional transform randomizes the point masses of $D_i$ via the uniform random variable $V\sim U(0,1)$. If $D_i$ is a continuous random variable, then $\tilde{F}_{D_i}(\cdot) = F_{D_i}(\cdot)$.

For the reminder of the section, let $D_i$ take values $t_k$ with probability $\P(D_i \le t_k) = p_k$, for $k = 1, \ldots, K$, such that $0=: p_0<  p_1 < \cdots < p_K :=1$. For $\tilde{U}$ defined in \eqref{eq:Di-in-terms-of-U-discrete} it holds $\P$-a.s. that
\begin{equation}\label{eq:D-discrete}
        D_i 
        = \sum_{k = 1}^{K-1} \Delta t_k\, \Id_{\{\tilde{U} \le p_k\}}  + t_K\,,
    \end{equation}
where $\Delta t_k := t_k - t_{k+1}$, $k = 1, \ldots, K-1$. Similar to \Cref{sec:bouded-support}, we perturb the latent standard normal variable that generates $D_i$, yielding
\begin{equation}\label{eq:D-discrete-pert}
        D_{i,\delta} 
    = \sum_{k = 1}^{K-1} \Delta t_k\, \Id_{\big\{ \Phi\big(\Phi^{-1}(\tilde{U})(1+ \delta)\big) \le p_k\big\}}
    + t_K\,.    
\end{equation}
The decision variable and its perturbations have then representation
    \begin{equation}\label{eq:Y-discrete-pert}
        Y 
        =
        \sum_{k = 1}^{K-1} \Delta_k \mfg\, \Id_{\{ \tilde{U} \le p_k\}}
        + \mfg_K
        \qquad \text{and} \qquad
        Y_{\delta} = \sum_{k = 1}^{K-1} \Delta_k \mfg\, \Id_{\{ \Phi\left(\Phi^{-1}(\tilde{U}) (1+ \delta)\right)\le p_k\}}  + \mfg_K
        \,,
    \end{equation}    
    where $ \Delta_k \mfg: = \mfg( \D_{-i},t_k, \X) - \mfg( \D_{-i},t_{k+1}, \X)$, $k = 1, \ldots, K -1$, $ \mfg_K := \mfg( \D_{-i},t_K, \X)$, and we use the notation $\D_{-i}:= (D_1, \ldots, D_{i-1}, D_{i+1}, \ldots, D_m)$, that is the vector $\D$, deprived of its $i$-th component. 
The next example shows how a Bernoulli rv, such as gender, is distorted.

\begin{example}[Bernoulli rv]\label{ex:bernoulli}
When $D\sim Ber(p)$ is a Bernoulli rv, i.e. $K = 2$, $p_1= 1-p$, $t_1 = 0$, $t_2 = 1$, we have
\begin{equation*}
D 
 =
    \begin{cases}
        1 \qquad  p
        \\
        0 \qquad  1-p
    \end{cases}
    \quad \text{and} \qquad
    D_\delta =
    \begin{cases}
        1 \qquad  p_\delta
        \\
        0 \qquad  1-p_\delta
    \end{cases},   
\end{equation*}
where $p_\delta:= 1 - \Phi\big(\frac{\Phi^{-1}(1-p)}{1 + \delta} \big)$. Thus, the perturbation is on the distributional leaving the values of $D$ fixed. Note that $\lim_{\delta \to 0}p_\delta = p$, for all $p \in [0,1]$. 
\end{example}

We now derive the sensitivity of the expected value and distortion risk measures under discrete perturbations.
\begin{proposition}\label{prop:sens-discrete-mean}
    Let $D_i$ take values $t_k$ with probability $\P(D_i \le t_k) = p_k$, for $k = 1, \ldots, K$ such that $0:= p_0<  p_1 < \cdots < p_K :=1$. Then the sensitivity to the expected value with perturbation given in \eqref{eq:D-discrete-pert}, is $\P$-a.s.
    \begin{equation*}
        \partial_{D_i} \, \E[Y|\X]
        =
        \sum_{k = 1}^{K-1} v_k \, \E \big[\, \Id_{\{D_i = t_k\}} \, \Delta_k \mfg\, |\,\X \big]\,,
\end{equation*}
where $v_k:=  -\Phi^{-1}(p_k) \,\phi\big(\Phi^{-1}(p_k)\big)$ for $k = 1, \ldots, K-1$.
\end{proposition}

This result states that the sensitivity to a discrete protected attribute $D_i$ depends on two key factors:  
(i) the conditional probability of each possible value $t_k$ of $D_i$ given the features $\X$; and  
(ii) how much the prediction function $\mfg$ changes when $D_i$ transitions from $t_{k+1}$ to $t_k$.
The coefficient $v_k$ represents how much the latent perturbation shifts probability mass around the threshold $p_k$, i.e., how ``perturbable'' the distribution of $D_i$ is at that level.

\begin{theorem}[Sensitivity - discrete]\label{thm:sens-discrete}
Let $D_i$ take values $t_k$ with probability $\P(D_i \le t_k) = p_k$, for $k = 1, \ldots, K$ such that $0:= p_0<  p_1 < \cdots < p_K :=1$. Then the sensitivity to a distortion risk measure $\rho_\gamma$ with perturbation given in \eqref{eq:D-discrete-pert}, is $\P$-a.s.
\begin{equation*}
    \partial_{D_i}\;  \rho_\gamma(\, Y\,|\,\X\,)
    =
    \sum_{k = 1}^K v_k\,\E\left[\Delta_k \mfg  \, \Id_{\{D_i = t_k\}} \gamma(U_{Y|\X})\, |\, \X = \x\right]\,,
\end{equation*}
where $v_k:=  -\Phi^{-1}(p_k) \,\phi\big(\Phi^{-1}(p_k)\big)$ for $k = 1, \ldots, K-1$, are the same as in \Cref{prop:sens-discrete-mean}.
\end{theorem}

Building on the intuition from the expected value risk measure case, the sensitivity of a distortion risk measure $\rho_\gamma(Y \mid \X)$ to a discrete protected attribute $D_i$ incorporates an additional factor: the distortion weight $\gamma(U_Y)$, where $U_Y$ denotes the rank of the outcome $Y$ within its conditional distribution. This weight modifies the contribution of each outcome based on its relative risk level—larger outcomes (e.g., high losses or costs) receive greater emphasis under risk-averse distortions. Consequently, the term $\Delta_k \mfg \cdot \gamma(U_Y)$ reflects not just how the outcome $Y$ changes across values of $D_i$, but also how important that outcome is from a risk-management perspective.

With the sensitivity result for discrete covariates in place, marginal fairness decision rules admit the same structural characterization as in the case of continuous covariates.

\begin{corollary}[Marginally fair decision rule]\label{cor:marginal-discrete}
    Let $D_i$ take values $t_k$ with probability $\P(D_i \le t_k) = p_k$, for $k = 1, \ldots, K$ such that $0:= p_0<  p_1 < \cdots < p_K :=1$. Consider the perturbation \eqref{eq:D-discrete-pert}, then 
    \begin{enumerate}[label = $\roman*)$]
        \item the marginal fair decision rule is given in \Cref{thm:individual}\,,
        \item the multi-marginal fair decision rule is given in \Cref{prop:mutli-marginal}, where each summand $l\in\{1, \ldots, m\}$ in \eqref{eq:muli-marginal-cont} is replaced by 
        $$
        \eta_l(\x)\, \sum_{k = 1}^K v_k\,\E\left[Y\, \Delta_k \mfg  \, \Id_{\{D_l = t_k\}} \, |\, \X = \x
    \right]\,,
        $$
    \end{enumerate}
 and where the sensitivity to $D_i$ is given in \Cref{thm:sens-discrete}.
    \end{corollary}

Next, we illustrate the sensitivity and the marginally fair conditional expectation

\begin{example}\label{ex:discrete-expected-value}

    We continue \Cref{ex:mean-linear} with $D$ taking values $t_k$ with probability $\P(D_i \le t_k) = p_k$, for $k = 1, \ldots, K$ such that $0:= p_0<  p_1 < \cdots < p_K :=1$. The sensitivity to $D$ is (note that $\Delta_k \mfg = \beta_2 (t_k - t_{k+1})$)
        \begin{equation}\label{eq:sens-discrete-regression}
        \partial_{D} \, \E[Y\,|\,X = x]
        =
           \beta_2\; \sum_{k = 1}^{K-1} v_{k}\, (t_k - t_{k+1} )\P(D = t_k | X = x)
\end{equation}
and the marginally fair premium is given in \eqref{eq:fair-decision-regression} with the sensitivity to $D_i$ in the coefficients replaced by \eqref{eq:sens-discrete-regression}.

  If further $D \sim Ber(p)$ as in \Cref{ex:bernoulli}, then the sensitivity to $D$ is
    \begin{equation}\label{eq:sens-bernoulli}
        \partial_{D_i} \, \E[Y|X = x]
        =
           \beta_2\, \Phi^{-1}(1-p) \,\phi\big(\Phi^{-1}(1-p)\big)\, \P(D = 0 | X = x)\,.
    \end{equation}
    For the special case when $D$ and $X$ are independent, i.e., $\P(D = 0 | X = x) = (1-p)$, the sensitivity as a function of the Bernoulli parameter $p$ is displayed in \Cref{fig:sens-bernoulli}. We observe that the sensitivity is larger/smallest for values of $p$ around 0.15 and 0.7, respectively. 

    \begin{figure}[th!]
        \centering
        \includegraphics[width=0.4\linewidth]{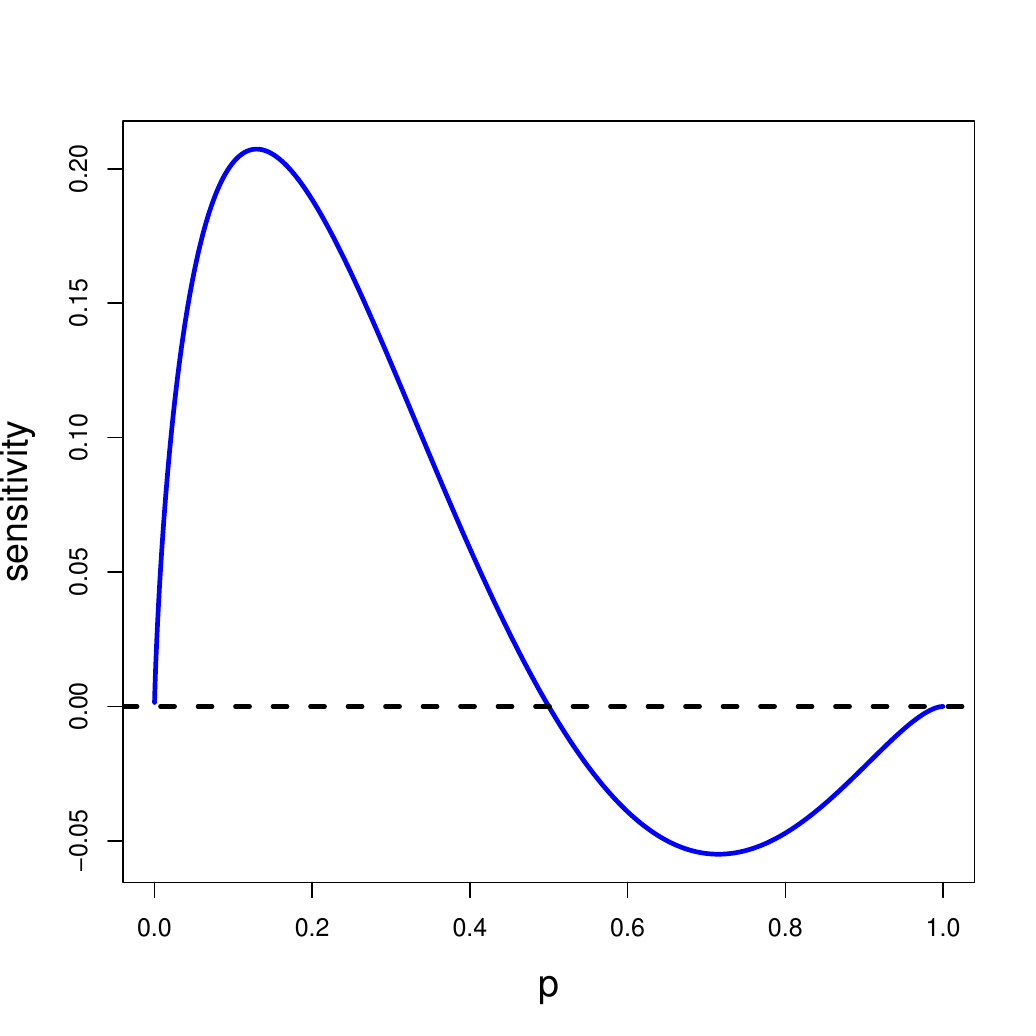}
        \caption{Sensitivity to $D$ under the assumption that $D\sim Ber(p)$ and independent of $X$ from \Cref{ex:discrete-expected-value}. The sensitivity is given in \eqref{eq:sens-bernoulli} with $\beta_2 = 1$. The $x$-axis is the success rate, i.e. $p \in (0,1)$.}
        \label{fig:sens-bernoulli}
    \end{figure}
\end{example}

\section{Marginal fairness with  cascade sensitivity}\label{sec:cascade}
 
In the presence of statistical dependence between the protected attributes \(\D\) and the remaining covariates \(\X\), a perturbation of a single protected feature \(D_i\) may influence other components of the input vector through their joint distribution. To account for this dependence structure, we extend the marginal fairness framework by introducing a \emph{cascade sensitivity} approach, in which perturbations propagate through the covariates via their joint copula. This construction allows for a fair assessment of decision rules under the assumption that the data-generating process is characterized by dependence without requiring causal assumptions. The representation via the Inverse Rosenblatt transform provides a probabilistic framework for generating perturbations consistent with the joint distribution, and forms the basis of the cascade perturbation defined below.

This cascade-based extension is particularly valuable for further addressing indirect discrimination, which arises when non-protected covariates serve as proxies for sensitive attributes due to their statistical dependence. Traditional fairness interventions that perturb protected features in isolation may underestimate the downstream effects of such dependencies. By modeling how perturbations propagate through the joint distribution of covariates, the cascade sensitivity approach ensures that fairness evaluations reflect the realistic structure of the data-generating process. This leads to a more robust and accurate assessment of whether a decision remains insensitive to both direct and indirect influences of protected attributes, even when those attributes are not explicitly used in the decision function.

If $(\D, \X)$ are dependent, then a perturbation on $D_i$ should cascade through the vector of covariates and change all other factors, i.e. $(\D_{-i}, \X)$, according to their statistical dependence with $D_i$. We consider the copula of $(\D, \X)$ which characterises statistical dependence and which does not require causal assumptions. By the Inverse Rosenblatt transform, it holds 
\begin{equation}\label{inv-Rosenblatt}
    (\D, \X) = \Big(\Psi^{(1)} (D_i, \V), \ldots, \Psi^{(m+n)} (D_i, \V)\Big) \quad \P\text{-a.s.}
\end{equation}
for some functions $\bPsi^{(k)} \colon \R^{m+n}\to \R$, $k \in \{1, \ldots, m+n\}$, and where $\V:= (V_1, \ldots, V_{m+n-1})$ are independent and identically distributed (i.i.d.) standard uniform rvs, and independent of $D$. Utilising representation \eqref{inv-Rosenblatt}, a perturbation on $D_i$ leads to the perturbed vector of covariates 
\begin{equation}\label{eq:pert-cascade}
    ( \D, \X)_\delta:= \big(\Psi^{(1)} (D_{i,\delta}, \V), \ldots, \Psi^{(m+n)} (D_{i,\delta}, \V)\big)\,.
\end{equation} 

To illustrate, consider 3 covariates $(D, X_1, X_2)$, in which case the standard construction of the inverse Rosenblatt transform becomes $\P$-a.s.
\begin{equation*} 
(D, X_1, X_2) = \Big(D, \,F^{-1}_{X_1|D}(V_1|D),\; F^{-1}_{X_2|X_1, D}(V_2 | X_1, D )\Big)   \,. 
\end{equation*} 
In this case the cascade perturbation becomes
\begin{equation*}
    (D, X_1, X_2)_\delta = \Big(D_\delta, \,F^{-1}_{X_1|D}(V_1|D_{\delta}),\; F^{-1}_{X_2|X_1, D}(V_2 | X_1 , D_{\delta}) \Big) \,,
\end{equation*}
thus all covariates are perturbed according to their statistical dependence with $D$.

The next example illustrates how one discrete protected covariate perturbs a non-protected covariate.
\begin{example}\label{mortgage-cascade}
 Let $Y \in \{0,1\}$ denote a mortgage decision ($Y = 1$ for approval), $X$ denote income, and $D \sim \text{Bern}(p)$ denote gender ($D = 1$ male, $D = 0$ female). Let $X \,|\, D = k \sim \log N\left((k+1)\mu, \sigma^2\right)$ for $k \in \{0,1\}$. Thus, on average, males earn twice as much as females. Next we impose the perturbation on $D$ given \Cref{ex:bernoulli}. Using the inverse Rosenblatt representation the cascade perturbation on $X$ becomes (see \Cref{app:ex}). 
\begin{equation*}
    X_\delta=
    \begin{cases}
    LogN(\mu, \sigma) \qquad &\text{with probability} \quad 1-p_\delta,\\
    LogN(2\mu, \sigma)  &\text{with probability} \quad p_\delta\\
    \end{cases}\,,
\end{equation*}
where $p_\delta$ is given in \Cref{ex:bernoulli}. \Cref{fig:cascade-preturbation} depicts the density of $X_\delta$ for $p  = 0.8$ (blue) and $p =0.2$ (red) and for $\delta = 0$ (solid) and for $\delta = 0.2$ (dashed).

    \begin{figure}
        \centering
        \includegraphics[width=0.4\linewidth]{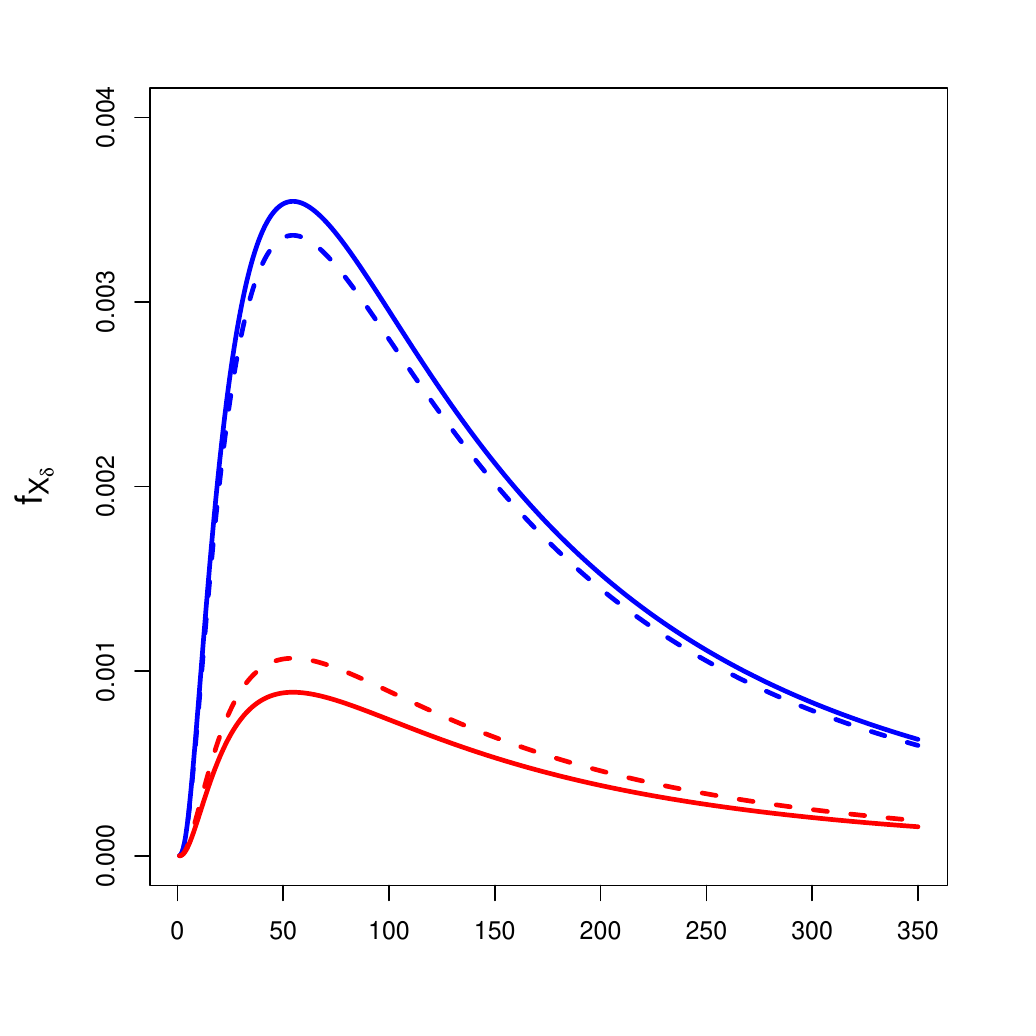}
        \caption{Cascading perturbation of $X_\delta$ due to a perturbation of $D \sim Bern(p)$ from \Cref{mortgage-cascade}. Blue lines correspond to $p = 0.8$ and red lines to $p =0.2$. Solid lines are $\delta = 0$ and dashed lines are the perturbation $\delta = 0.2$.}
        \label{fig:cascade-preturbation}
    \end{figure}
\end{example}

Next, we define the cascade sensitivity. 

\begin{definition}[Cascade sensitivity]
    Let $\supp(D_i) = \R$ and consider the perturbation defined in \eqref{eq:pert-cascade}. Then the cascade sensitivity is 
\begin{equation}
   \partial^c_{D_i}\;  \rho_\gamma(\, Y\,|\,\X\,) 
   :=
   \lim_{\delta \downarrow 0}\frac{\rho_\gamma\big(\mfg\big((\D, \X)_\delta\big)\big) - \rho_\gamma\big(\mfg(\D, \X)\big)}{\delta}.
\end{equation}
\end{definition}

We say that a decision rule $\rho_\gamma$ is \textit{marginally fair with cascade sensitivity}, if the cascade sensitivity to covariate $D_i$ vanishes. Similarly, we say the decision rule is \textit{multi-marginally fair with cascade sensitivity}, if the decision rule is marginally fair with cascade sensitivity for all protected covariates $D_i$, $i = 1, \ldots, m$. We first establish a representation of the cascade sensitivity, and second show that the main \Cref{thm:individual} (in slight modification) still applies.

We first provide the sensitivity formulas to protected covariates whose support is the real line and then state the corresponding results for compactly supported and discrete covariates.

\begin{theorem}[Cascade sensitivity]\label{thm:casaade-sensitivity}
Let $\supp(D_i) = \R$ and consider the perturbation $D_{i,\delta}: = D_i (1 + \delta)$. Assume that $\mfg$ is invertible in its $i$-th component and that for all $u \in (0,1)$, the function $\delta \to \Finv_{\mfg\big( (\D, \X)_\delta \big)}(u)$, $\delta \ge 0$, is differentiable in a neighborhood of $\delta = 0$ with bounded derivative. Then the cascade sensitivity has $\P$-a.s. representation 
\begin{align*}
   \partial^c_{D_i}\;  \rho_\gamma(\, Y\,|\,\X\,)
        &=
         \E\big[\,D_i \,\partial_{i} \mfg(\D, \X) \gamma\big(U_{Y|\X}\big)  \,|\, \X\,\big]
        \\
        & \quad 
        +
       \sum_{\substack{l = 1 \\ l \neq i}}^{m+n} \E\big[\,D_i \,\partial_{l} \mfg(\D, \X) \big(\tfrac{\partial}{\partial t}F^{-1}_{l|D_i}(V|t)\big|_{t = D_i}\big) \gamma\big(U_{Y|\X}\big)  \,|\, \X\,\big]\,,
    \end{align*}
    where $V\sim U(0,1)$ independent of $(\D,\X)$,$F_{l|D_i}(\cdot|d):= \P(D_l \le \cdot |D_i = d) $ for $l \in\{ 1, \ldots, n\}$, and $F_{j|D_i}(\cdot|d):= \P(X_j \le \cdot |D_i = d) $ for $j = n+1, \ldots, n+m$.
\end{theorem}
We observe that the cascade sensitivity is composed of the sensitivity to $D_i$ (which is the summand with $l = i$ as $F_{i|D_i}(\cdot|D_i):= \P(D_i \le  \cdot |D_i)$ is the identity), and $m+n-1$ summands each of which correspond to how much the covariates $D_k$, $k \neq i$, and $X_j$ are contributing to the sensitivity of $D_i$. The summand $l = n+j$, for example, corresponds to the sensitivity to the indirect perturbation of $X_j$ (via $D_i$) and the perturbation on $X_j$ is captured through the term $\tfrac{\partial}{\partial t}F^{-1}_{l|D_i}(V|t)\big|_{t = D_i}$.  Marginal fairness with cascade sensitivity mitigates indirect discrimination through accounting both paths $\D \rightarrow Y \rightarrow \rho_\gamma(Y|\X)$ and $\D \rightarrow \X \rightarrow Y \rightarrow \rho_\gamma(Y|\X)$ in \Cref{fig:decision-graph}. 

\begin{proposition}[Marginally fair decision rule]\label{prop:marginal-cascade}
    Let $\supp(D_i) = \R$ and consider the perturbation \eqref{inv-Rosenblatt} with $D_{i, \delta}: = D_i(1 + \delta)$. Then 
    \begin{enumerate}[label = $\roman*)$]
        \item the marginally fair decision rule with the cascade sensitivity is given in \Cref{thm:individual}\,,
        \item the multi-marginal fair decision rule with the cascade sensitivity is given in \Cref{prop:mutli-marginal}, where each summand $l \in \{1, \ldots, m\}$ in \eqref{eq:muli-marginal-cont} is replaced by 
        $$\eta_l(\x)\sum_{k = 1 }^{m+n} \E\big[\,Y\, D_l \,\partial_{k} \mfg(\D, \X) \big(\tfrac{\partial}{\partial t}F^{-1}_{k|D_l}(V|t)\big|_{t = D_l}\big)  \,|\, \X\,\big], $$
    \end{enumerate}
    and where the sensitivity, $\partial_{D_i}\;  \rho_\gamma(\, Y\,|\,\X\,),$ is replace with the cascade sensitivity $\partial^c_{D_i}\;  \rho_\gamma(\, Y\,|\,\X\,)$ given in \Cref{thm:casaade-sensitivity}. 
\end{proposition}

The cascade sensitivity is fundamentally based on statistical dependence, i.e. the copula, between the covariates $(\D, \X)$ without making any causal assumptions. If the decision maker, however, has access to a causal graph or a partial causal graph of $(\D, \X)$, this information can be integrated into the cascade sensitivity. Indeed, the cascade perturbation assumes that a change in, say $D_i$, affects all other covariates $(\D_{-i}, \X)$. With a causal graph stating that, e.g., $D_k, X_j$, $k \in \mcK \subseteq\{1, \ldots, m\}$, $j \in \mcJ \subseteq\{1, \ldots, n\}$, are (indirectly) causing $D_i$, then of course a perturbation of $D_i$ should leave $D_k, X_j$, $k \in \mcK, j \in \mcJ$ unaffected. Thus, in this situation, the decision maker considers the cascade sensitivity in \Cref{thm:casaade-sensitivity} given by 
\begin{align*}
       \partial^c_{D_i}\;  \rho_\gamma(\, Y\,|\,\X\,)
        &=
       \sum_{\substack{l = 1,  \\ l \notin \mcK, \; l \notin \mcJ}}^{m+n} \E\big[\,D_i \,\partial_{l} \mfg(\D, \X) \big(\tfrac{\partial}{\partial t}F^{-1}_{l|D_i}(V|t)\big|_{t = D_i}\big) \gamma\big(U_{Y|\X}\big)  \,|\, \X\,\big]\,,
\end{align*}
that is, the summands corresponding to the indirect sensitivities of  $D_k, X_j$, $k \in \mcK, j \in \mcJ$ are removed.

Next, we provide an example of the marginally fair expected value with cascade sensitivity.
\begin{example}\label{ex:cascade32}
    We continue \Cref{ex:mean-linear} with $\supp(D) = \R$. Then the cascade sensitivity for the conditional expectation to $D$ is
\begin{align*}
    \partial^c_{D}\;  \rho_1(\, Y\,|\,X = x\,)
        &=
        \partial_{D}\;  \rho_1(\, Y\,|\,X = x\,)
       +
       \beta_1\, \E\big[\,D \,\big(\tfrac{\partial}{\partial t}F^{-1}_{X|D}(V|t)\big|_{t = D}\big) \,|\, X = x\,\big]
\end{align*}
and the marginally fair premium becomes
    \begin{align*}
        \rho^{D}_{1*}(Y \,|\,X = x) 
        &=
    \beta^\dagger_0(x) + \beta_1^\dagger(x)\, x \,,
\end{align*} 
where the coefficients are $\beta^\dagger_0(x): = \tilde{\beta}_0(x)  - \beta_0 \beta_1(1 - \frac{c_x}{\beta_2} )H(x)$, $\beta^\dagger_1(x): = \tilde{\beta}_1(x) + \frac{\beta_1^2}{\beta_2}c_x H(x)$, with $H(x): =\E\big[\,D \,\big(\tfrac{\partial}{\partial t}F^{-1}_{X|D}(V|t)\big|_{t = D}\big) \,|\, X = x\,\big]$,
and $\tilde{\beta}_0(x)$, $\tilde{\beta}_1(x)$, and $c_x$ given in \Cref{ex:mean-linear-fair}.
\end{example}

\subsection{Cascade sensitivity for compactly supported, discrete, and categorical covariates}
The cascade sensitivity also applies to protected covariates that are continuously distributed with compact support, that are discrete, or categorical.

\begin{proposition}[Cascade sensitivity - compact support]\label{prop:cascade-compact}
Let $D_i$ be an absolutely continuous rv with compact support and the perturbation given in \eqref{eq:perturb-compact}. Assume that $\mfg$ is invertible in its $i$-th component and that for all $u \in (0,1)$, the function $\delta \to \Finv_{\mfg\big( (\D, \X)_\delta  \big)}(u)$ is differentiable in a neighborhood of $\delta = 0$ with bounded derivative.
Then the cascade sensitivity has $\P$-a.s. representation 
\begin{align*}
   \partial^c_{D_i}\;  \rho_\gamma(\, Y\,|\,\X\,)
        &=
       \sum_{\substack{l = 1 }}^{m+n} \E\Big[\,\frac{\phi\big(\Phi^{-1}(F_{D_i}(D_i))\big)}{f_{d_i}(D_i)} \,\partial_{l} \mfg(\D, \X) \big(\tfrac{\partial}{\partial t}F^{-1}_{l|D_i}(V|t)\big|_{t = D_i}\big) \gamma\big(U_{Y|\X}\big)  \,|\, \X\,\Big]\,.
    \end{align*}
\end{proposition}

\begin{proposition}[Cascade sensitivity - discrete]\label{prop:cascade-discrete}
Let $D_i$ take values $t_k$ with probability $\P(D_i \le t_k) = p_k$, for $k = 1, \ldots, K$ such that $0:= p_0<  p_1 < \cdots < p_K :=1$. Then the cascade sensitivity to a distortion risk measure $\rho_\gamma$ with perturbation given in \eqref{eq:D-discrete-pert}, is $\P$-a.s.
\begin{equation*}
    \partial_{D_i}^c\;  \rho_\gamma(\, Y\,|\,\X\,)
    =
    \sum_{l = 1}^{n+m}\, 
    \sum_{k = 1}^{K-1}v_k\, 
    \E[\Delta_{k,l} \tilde{\mfg} \, \Id_{\{D_i = t_k\}} \,  \gamma(U_{Y|\X})\,| \X = \x]\,,
\end{equation*}
where $ \Delta_{k,l} \tilde{\mfg}: =
\mfg\big((\D, \X)_{-l}, F^{-1}_{X_l}(V\mid D_i=t_k)\big) - \mfg\big((\D, \X)_{-l}, F^{-1}_{X_l}(V\mid D_i=t_{k+1})\big)$, $k = 1, \ldots, K -1$, and $ \tilde{\mfg}_{K,l} := \mfg\big((\D, \X)_{-l}, F^{-1}_{X_l}(V\mid D_i=t_K)\big)$, and $l = 1\ldots, m+n$. 
\end{proposition}

Furthermore, the marginal sensitivity of \Cref{cor:marginal-compact} (\Cref{cor:marginal-discrete}) holds with the assumptions replaced by the assumptions of \Cref{prop:cascade-compact} (\Cref{prop:cascade-discrete}) and the sensitivities replace by the respective cascade sensitivities established in this section. For the multi-marginal sensitivities for protected covariate with compact support, each summand $l\in\{1, \ldots, m\}$  in \eqref{eq:muli-marginal-cont} needs replaced by 
$$
\eta_l(\x) \sum_{\substack{k = 1 }}^{m+n} \E\Big[\,Y\, \frac{\phi\big(\Phi^{-1}(F_{D_l}(D_l))\big)}{f_{d_l}(D_l)} \,\partial_{k} \mfg(\D, \X) \big(\tfrac{\partial}{\partial t}F^{-1}_{k|D_l}(V|t)\big|_{t = D_l}\big)   \,|\, \X\,\Big]\,.
$$
Similarly, for discrete protected covariates, each summand in \eqref{eq:muli-marginal-cont} needs to be replaced by 
$$
\eta_l(\x)\sum_{r = 1}^{n+m}\, 
    \sum_{k = 1}^{K-1}v_k\, 
    \E[Y \Delta_{k,r} \tilde{\mfg} \, \Id_{\{D_l = t_k\}}\,| \X = \x]\,.
$$

\section{Numerical study} \label{sec:simulation} 
In this section, we illustrate the impact of enforcing marginal fairness on decision-making through a numerical study. We highlight how traditional fair decision rules compare to decision rules adjusted for marginal fairness under both expected value (EV) and ES.  We assume a data-generating process in which the non-protected feature $X$ and the protected attribute $D$ follow a bivariate normal distribution:
\begin{equation*}
    \begin{bmatrix} X \\ D \end{bmatrix} \sim \mathcal{N}\left(
    \begin{bmatrix} \mu_X \\ \mu_D \end{bmatrix},
    \begin{bmatrix} \sigma_X^2 & \tau \sigma_X \sigma_D \\ \tau \sigma_X \sigma_D & \sigma_D^2 \end{bmatrix}
    \right),
\end{equation*}
with parameters $\mu_X = 0$, $\mu_D = 3$, $\sigma_X = 1$, $\sigma_D = 2$, and correlation $\tau = 0.5$. The response variable \(Y\) is generated via a linear model:
\begin{equation*}
    Y = \beta_0 + \beta_X X + \beta_D D + \varepsilon,
\end{equation*}
where \( \varepsilon \sim \mathcal{N}(0, \sigma_\varepsilon^2) \) with \( \sigma_\varepsilon = 0.5 \), and coefficients $\beta_0 = 1$, $\beta_X = 2$, $\beta_D = 1$.

This setup mirrors settings where protected, non-protected, and outcome variables are treated as continuous. The protected attribute $D$ may represent credit score, age, or a socioeconomic index---features that, while continuous, are often regulated or restricted in decision-making due to fairness concerns. The non-protected attribute $X$ may correspond to vehicle value, annual mileage, or income, which are commonly used in insurance pricing and financial risk assessment. The response variable $Y$ reflects a continuous outcome such as insurance claims, loan default loss, or healthcare expenditure. 

Although we assume joint normality for $(D, X)$, our framework does not rely on this distributional assumption and readily applies to settings with non-Gaussian covariates or outcomes. Moreover, variables that are strictly positive and exhibit skewness in practice can often be transformed (e.g., via logarithmic or power transformations) to approximate normality. In the empirical study in \Cref{sec:empirical}, we adopt alternative loss functions that better reflect real-world data characteristics. For this section, we focus on simplified distributions to ensure transparency and tractability in illustrating the effects of fairness adjustments.

\subsection{Marginal fairness with marginal sensitivity}

We consider four decision strategies for a fixed feature value  \( X=x \) below:
\begin{enumerate}[label = $(\roman*)$]
    \item \textit{Unaware decision:} Removing 
$D$ from decision-making process, that is
    \begin{equation*}
        P_U = \E[Y | X]=  \beta_0 + \beta_X x + \beta_D \mathbb{E}[D | X=x].
    \end{equation*}
    \item \textit{Discrimination-free decision:} Removing proxy discrimination by averaging out the protected attribute $D$ \cite{Lindholm2022ASTIN,pope2011implementing}
    \begin{equation*}
        P_{DF} = \beta_0 + \beta_X x + \beta_D \mathbb{E}[D].
    \end{equation*}
    \item \textit{Marginally fair decision with EV:} Adjusted to achieve marginal fairness with marginal sensitivity for the expected value decision rule
    \begin{equation*}
        P_{MF_{EV}} = \beta_0  (1-c_x) + \beta_X  (1-c_x)  x,
    \end{equation*}
   where the adjustment factor $c_x$ depends on moments of $D$ conditional on $X$, as defined in \Cref{ex:mean-linear-fair}. 
    \item \textit{Marginally fair decision with ES:}  Adjusted to achieve marginal fairness with marginal sensitivity for ES at $\alpha=0.95$,
    \begin{equation*}
        P_{MF_{ES}} = \bar{\beta}_0(x)  + \bar{\beta}_X(x)   x,
    \end{equation*}
   where $\bar{\beta}_0(x)$ and $\bar{\beta}_1(x)$ are given in \Cref{ex:mean-linear-fair} with $\gamma(u) = \frac{1}{1-\alpha} \Id_{\{u \ge \alpha\}} $.

\end{enumerate}
The four decision strategies differ fundamentally in how they achieve fairness. 
The \emph{unaware decision} enforces fairness through exclusion, omitting the protected attribute $D$ entirely from the decision rule. 
The \emph{discrimination-free decision} achieves fairness by averaging out $D$ across the population, while maintaining the original coefficient on $X$. 
In contrast, the two \emph{marginally fair decisions}---under expected value and ES---achieve fairness by adjusting the coefficients themselves. 
These adjustments are specifically designed to eliminate the derivative-based sensitivity of the decision rule with respect to $D$, ensuring that decisions are insensitive to small perturbations in the protected attribute.

\begin{figure}[th]
    \centering
    \includegraphics[width=0.6\textwidth]{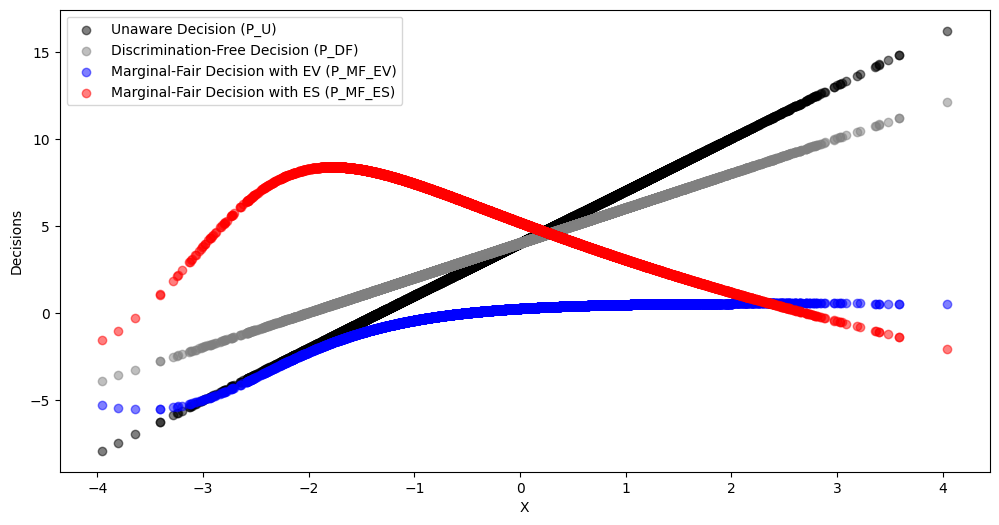}  
    \caption{Comparison of fair decision strategies. Unaware decision with expected value (black), discrimination-free with expected value (grey), and marginally fair with expected value (blue) and ES (red).}
    \label{fig: com-sim}
\end{figure}

\Cref{fig: com-sim} compares the four decision strategies across a range of values for the non-protected attribute $X$. The unaware and discrimination-free decision rules both produce linear relationships between $X$ and the predicted decision, with differing slopes reflecting their treatments of the protected attribute $D$. In contrast, both marginally fair decisions introduce nonlinear adjustments to the decision rule. These adjustments are more pronounced in regions where the influence of $D$ on the risk measure applied to response variable $Y$ is stronger, illustrating how marginal fairness explicitly corrects for sensitivity to protected attributes. 

\begin{figure}[h]
    \centering
    \includegraphics[width=0.6\textwidth]{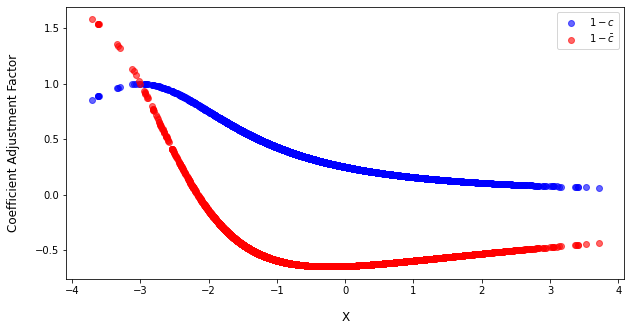}  
    \caption{Coefficient adjustment $1 - c_x$ and $1 - \bar{c}_x$ for marginally fair decisions under expected value and ES, respectively.}
    \label{fig: c-sim}
\end{figure}

\Cref{fig: c-sim} compares the coefficient adjustment factors \(1 - c_x\) and \(1 - \bar{c}_x\), which correspond to marginal fairness for expected value and ES, respectively. Both adjustment factors vary with \(X\), highlighting that fairness corrections are data-driven rather than uniform. While both curves decline as \(X\) increases (for large enough $X$), the adjustment under ES exhibits sharper variation, particularly in the left tail of the distribution, while, the adjustment under expected value is smoother, suggesting a more uniform correction. The divergence between the two curves illustrates how the choice of fairness criterion influences the character and magnitude of decision adjustments.

\subsection{Marginal fairness with cascade sensitivity}
We then move on to apply cascade sensitivity to the marginally fair decision rule based on the expected value measure according to \Cref{ex:cascade32}. In this setting, the conditional distribution $X \mid D$ is Gaussian with mean $\mu_{X|D}(D) = \mu_X + \tau \frac{\sigma_X}{\sigma_D}(D - \mu_D)$ and constant variance $\sigma_{X|D}^2 = \sigma_X^2(1 - \tau^2)$. The conditional quantile function is thus $F^{-1}_{X|D}(V|t)\big|_{t = D} = \mu_{X|D}(t) + \sigma_{X|D} \Phi^{-1}(V)$, from which we obtain $\tfrac{\partial}{\partial t}F^{-1}_{X|D}(V|t)\big|_{t = D} = \tau \frac{\sigma_X}{\sigma_D}$ as a constant. The cascade sensitivity is then computed as
\[
\partial^c_D \mathbb{E}[Y \mid X = x] = \beta_2 \mathbb{E}[D \mid X = x] + \beta_1 \cdot \tau \frac{\sigma_X}{\sigma_D} \cdot \mathbb{E}[D \mid X = x].
\]
This expression combines the direct sensitivity of the decision rule to the protected attribute $D$ with the indirect effect captured via the copula-induced dependence of $X$ on $D$. We can then further derive the $H(x)$ in \Cref{ex:cascade32} as $H(x) = \tau \frac{\sigma_X}{\sigma_D} \cdot \mathbb{E}[D \mid X = x]$,
which can be computed analytically using the formula for the conditional expectation under joint normality, $\mathbb{E}[D \mid X = x] = \mu_D + \tau \frac{\sigma_D}{\sigma_X}(x - \mu_X)$.  Finally, we derive the marginal fair decision with cascade sensitivity 
and the marginally fair premium becomes
    \begin{align*}
       P^c_{{MF}_{EV}}
        &=
    \beta^\dagger_0(x) + \beta_1^\dagger(x)\, x \,,
\end{align*}  
where the coefficients are $\beta^\dagger_0(x)$ and $\beta^\dagger_1(x)$ are given in \Cref{ex:cascade32}.

\begin{figure}
    \centering
    \includegraphics[width=0.6\linewidth]{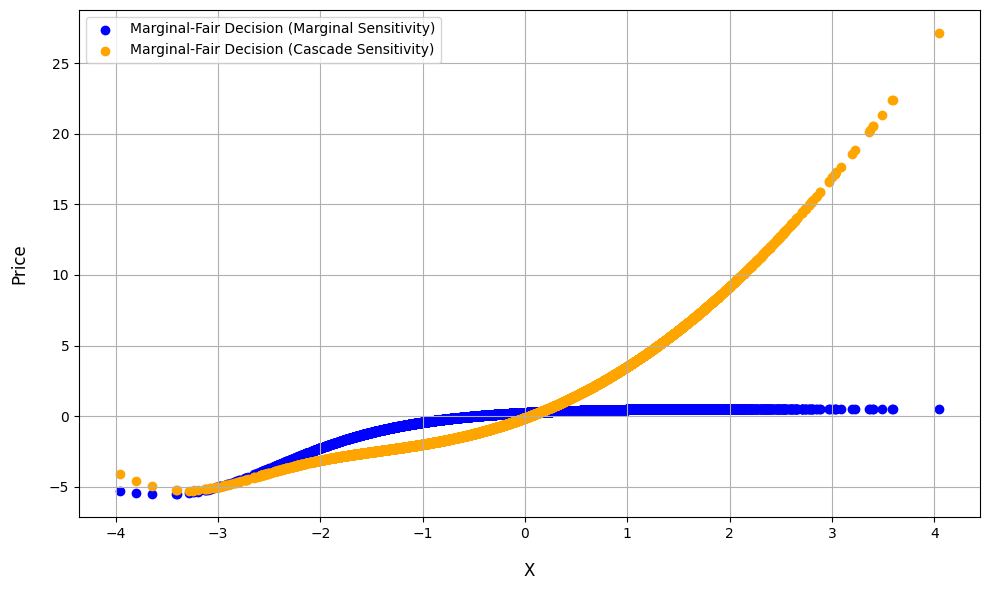}
    \caption{Comparison of fair decision strategies based on expected value. Marginally fair decision with marginal sensitivity (blue) and marginally fair decision with cascade sensitivity (orange).}
    \label{fig:comparison_fair_simulation}
\end{figure}

\begin{figure}
    \centering
    \includegraphics[width=0.6\linewidth]{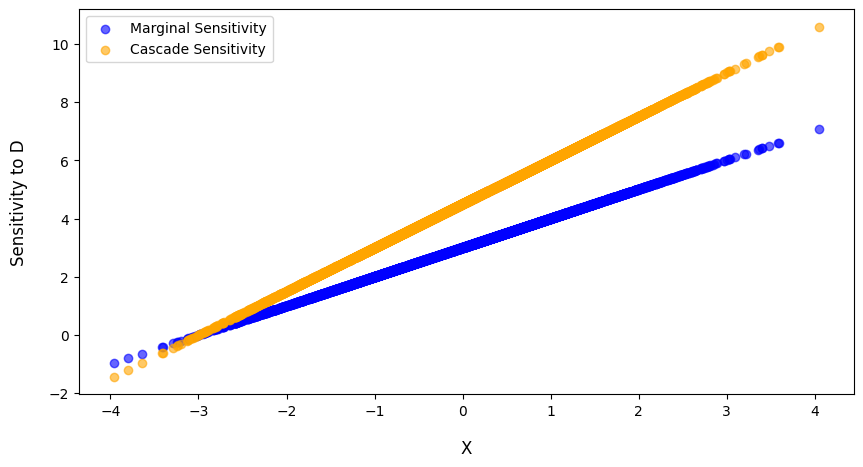}
    \caption{Comparison of sensitivity of marginally fair decisions with marginal sensitivity (blue) and cascade sensitivity (orange) based on expected value. }
    \label{fig:sim_sensitivity_cas}
\end{figure}

Figures~\ref{fig:comparison_fair_simulation} and~\ref{fig:sim_sensitivity_cas} illustrate the impact of cascade sensitivity in fair decision-making based on the expected value risk measure. In Figure~\ref{fig:comparison_fair_simulation}, we compare the marginally fair decision rules obtained under marginal sensitivity and cascade sensitivity. Both approaches adjust the original decision rule to ensure fairness, but the cascade-sensitive rule accounts for indirect effects of the protected attribute through its dependence with other covariates, resulting in a noticeable shift in decisions, particularly in the tails of the distribution. Figure~\ref{fig:sim_sensitivity_cas} visualizes the corresponding sensitivities. The marginal sensitivity remains moderate across the range of $X$, while the cascade sensitivity shows a larger and more variable effect due to the additional contribution from the dependence structure between $D$ and $X$. These results highlight that neglecting cascade effects may underestimate the influence of protected attributes on decisions, and that marginal fairness with cascade sensitivity provides a more accurate adjustment in such settings.

\section{Empirical implementation} \label{sec:empirical}
We complement our numerical study with an empirical case study based on real-world insurance data using the categorical protected variable gender. 
A key real-world motivation for our empirical study is the implementation of gender-neutral pricing regulations in insurance markets, most notably the European Union’s Gender Directive (Directive 2004/113/EC). This regulation prohibits insurers from using gender as a rating factor in determining premiums, even though gender may still be used in the modeling of claim costs. As a result, insurers face the challenge of producing fair and legally compliant pricing decisions while preserving predictive accuracy. 

For this application, we first assume that the decision rule is based on the expected value of the outcome. This choice facilitates direct comparison with benchmark pricing models that rely on mean-based predictions. We then implement the ES as an additional decision rule that focuses on tail risk. The marginally fair decision rule is applied based on marginal sensitivity, as the dependence between the protected variable ``gender'' and other non-protected variables is relatively weak in this dataset \cite{xin2024}. 


\subsection{Dataset}
We analyze a dataset (\texttt{pg15training}) from French private motor insurance, sourced from the \texttt{R} package \texttt{CASdatasets} \cite{dutang2020package}, which has been used in prior research \cite{xin2024, vincent2022fair}. It contains 100,000 third-party liability policies observed over four years.

The dataset includes key variables such as policyholder demographics, vehicle characteristics, claim frequency, and severity. Gender is assumed to be the protected variable. The total third-party claims cost is the response variable for claims cost modeling. We preprocess the dataset by removing irrelevant variables such as \texttt{CalYear}, \texttt{SubGroup2}, \texttt{Indtpbi}, \texttt{Numtpbi}, and \texttt{Bonus}, binning age groups into 10-year intervals for better interpretability and creating exposure-weighted variables for claims modeling. \Cref{tab:variables} summarizes the key variables used in our models. We split the dataset into 70\% for training and 30\% for testing. The training set is used to fit the model and estimate its parameters, while the testing set is used to evaluate model outputs for new, unseen customers. All empirical results are based on the out-of-sample testing set, using the model trained on the training set.

\begin{table}[h]
    \centering
    \small
    \caption{Variables used in the dataset}
    \begin{tabular}{ll}
        \toprule
        \textbf{Variable} & \textbf{Description} \\
        \midrule
        PolNum & Policy number \\
        Gender & Driver's gender \\
        Type & Car type (6 categories: A, B, C, D, E, F) \\
        Category & Car category (Large, Medium, Small) \\
        Occupation & Driver occupation (Employed, Housewife, Retired, etc.) \\
        Age & Driver's age (binned into groups) \\
        Group1 & Car classification (20 categories) \\
        Poldur & Policy duration (years) \\
        Value & Car value (euro) \\
        Adind & Additional voluntary cover (dummy variable) \\
        Group2 & Driver region (10 categories) \\
        Density & City population density \\
        Exppdays & Exposure in days \\
        Numtppd & Number of third-party claims \\
        Indtppd & Total third-party claim cost (euro) \\
        \bottomrule
    \end{tabular}
    \label{tab:variables}
\end{table}

\subsection{Modeling process}
We examine the problem of insurance pricing based on the expected value risk measure using a two-step decision-making process in this section. In the first step, we model the claims cost using available covariates. In the second step, we apply a generalized distortion risk measure to inform pricing decisions. For this application, we focus on the expected value as the decision rule to enable direct comparison with benchmark pricing models.

 Following common practice to model insurance claims \cite{goldburd2016generalized}, we fit a generalized linear model (GLM) with Tweedie loss to estimate the prediction function $Y=\mfg(\D, \X)$ using both $\D$ and $\X$. This step reflects best-practice predictive modeling in insurance pricing. Compared to black-box models, GLMs offer improved interpretability, which is important in regulated domains such as insurance.
To estimate the expected value risk measure \( \rho_1(Y \mid X)= \mathbb{E}[Y \mid \X] \), we fit a second GLM  with Tweedie loss using only non-protected variables $\X$.  This ensures that the decision rule is based solely on admissible information, avoiding direct discrimination. Both modeling steps are implemented in PyTorch using the Adam optimizer, which provides flexibility to increase model complexity if needed and supports gradient-based sensitivity analysis in subsequent stages of our fairness-aware learning framework. We have also implemented more complex models for both steps using neural networks (NNs), which yielded similar results. In this paper, we focus on presenting the GLM-based results to maintain consistency with standard practice in the insurance industry.

We then apply \Cref{cor:marginal-discrete} to calculate the marginally fair decision rule for the expected value as the protected covariate is discrete. To do so, we need to estimate four components. First (1) the derivative of the prediction function $\partial_{D_i} \mfg(D, X)$ is computed using automatic differentiation and the chain rule, as implemented within the PyTorch framework. (2) the sensitivity of the decision rule $\partial_{D_i} \rho_\gamma(Y \mid X)$ is given in \Cref{prop:sens-discrete-mean}.  
As $v_k$ and $\Delta_k \mfg$ are a constant conditional on $\X$ and $\P( D_i = t_k \,  |\,\X )$  is estimated using a NN with binary cross entropy loss to predict $\D_i$ using $\X$, the sensitivity of the decision rule can be readily computed.
(3) $\E\big[Y \, D_i \, \partial_{i} \mfg(\D, \X)\, |\,\X \big]$ is modeled using a NN with Tweedie loss to predict $Y \, D_i \, \partial_{i} \mfg(\D, \X)$ using $\X$, and (4) $\E\big[\,\big(D_i \partial_{i} \mfg(\D, \X) \big)^2 \big|\,\X\big]$ is modeled using a NN with gamma loss to predict $\big(D_i \partial_{i} \mfg(\D, \X) \big)^2$ using $\X$.

Model specifications are provided in the \Cref{app:model}. We then apply the marginal fairness correction formula from \Cref{cor:marginal-discrete} to obtain the marginally fair decision rule, which is directly comparable to the unaware and discrimination-free rules.

\subsection{Results and interpretation} \label{sec:Em_results}

This section presents the results and interpretation based on the expected value risk measure. Similar to \Cref{sec:simulation}, we compare our results with two benchmark decision rules: the unaware decision and the discrimination-free decision. The unaware decision rule, \(P_U\), is obtained without any fairness adjustments --- this is the common industry practice to address EU gender-neutral pricing regulation. The discrimination-free decision rule, \(P_{DF}\), is constructed by averaging out the protected attribute following the approach of \cite{Lindholm2022ASTIN,pope2011implementing}. For reference, we denote our proposed marginally fair decision rule by \(P_{MF}\). 

\Cref{tbl:summaryDecision} shows the summary statistics of the distribution of decisions under the three fairness criteria: marginally fair, discrimination-free, and unaware decision. For each rule, we present the minimum and maximum decision values, as well as the 25th, 50th (median), and 75th percentiles.  Among the three, the unaware decision exhibits the highest maximum and the highest quantiles across the board. The marginally fair decisions fall between the discrimination-free and unaware decisions in terms of magnitude.  A portfolio-level rebalancing property can be achieved using techniques proposed in \cite{Lindholm2022ASTIN}. 

\begin{table}[ht]
\centering
\scalebox{0.9}{
\begin{tabular}{lccccc}
\toprule
\textbf{Decision Rule} & \textbf{Min}  & \textbf{25\% Quantile} & \textbf{50\% Quantile} & \textbf{75\% Quantile} & \textbf{Max} \\
\midrule
Marginally Fair, $P_{MF}$        & 4.60  & 39.41  & 67.89  & 125.78 & 1506.90 \\
Discrimination-Free, $P_{DF}$    & 2.86  & 35.09  & 61.67  & 117.90 & 1557.90 \\
Unaware, $P_U$                   & 4.82  & 41.57  & 71.89  & 133.16 & 1599.57 \\
\bottomrule
\end{tabular}}
\caption{Summary statistics of decisions under different fairness criteria under the expected value risk measure}
\label{tbl:summaryDecision}
\end{table}

\Cref{tbl:summaryPU-PMF} presents summary statistics that quantify the central tendency and variability of the difference between the unaware and the marginally fair decision \( P_U - P_{MF} \). Note that this is exactly the adjustment term to make the expected value decision margainally fair, i.e. the second term in \eqref{eq:marginally-fair}. The values are strictly positive, reflecting the negative coefficient of $\D$ in the GLM model for the prediction function $Y = \mfg(\D, \X)$. This implies a negative sensitivity of the prediction function $\mfg()$ with respect to $\D$, indicating that claims costs are, on average, negatively associated with gender (i.e., being female is associated with lower expected claims). Consequently, both the sensitivity of the decision rule, \( \partial_{D_i} \rho_\gamma(Y \mid X) \), and \( \mathbb{E}\big[Y \, D_i \, \partial_{i} \mfg(\D, \X)\, \mid\,\X \big] \), take negative values.
\begin{table}[H]
\centering
\scalebox{0.9}{
\begin{tabular}{lccccc}
\toprule
\textbf{Statistic} & \textbf{Min} & \textbf{25\% Quantile} & \textbf{50\% Quantile} & \textbf{75\% Quantile} & \textbf{Max} \\
\midrule
$P_U - P_{MF}$     & 0.08         & 1.99                   & 3.76                   & 7.29                   & 115.23 \\
\bottomrule
\end{tabular}}
\caption{Summary statistics of the difference between unaware and marginally fair decisions, $P_U - P_{MF}$, under the expected value risk measure}
\label{tbl:summaryPU-PMF}
\end{table}

\begin{figure}
    \centering
    \includegraphics[width=0.6\linewidth]{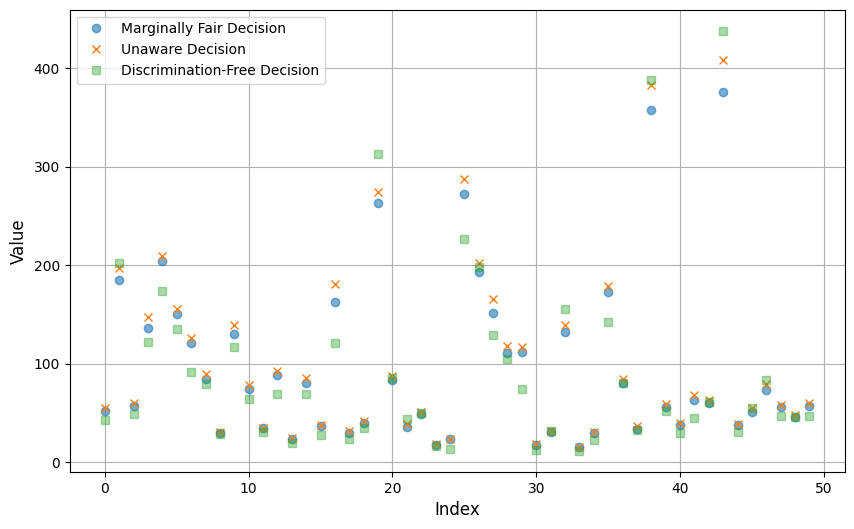}
    \caption{Comparison of fair decision strategies for 50 randomly selected policyholders under the expected value risk measure. Blue dots correspond to marginally fair decision, orange crosses to unaware decision, and the greed squares to the discrimination-free decision.}
    \label{fig:50points}
\end{figure}

\Cref{fig:50points} illustrates the three decision strategies for 50 randomly selected policyholders. Overall, the strategies yield similar decisions and are largely aligned. For some individuals, all three decision rules produce nearly identical outcomes, while for others, the differences are more pronounced.

\begin{figure}
    \centering
    \includegraphics[width=0.44\linewidth]{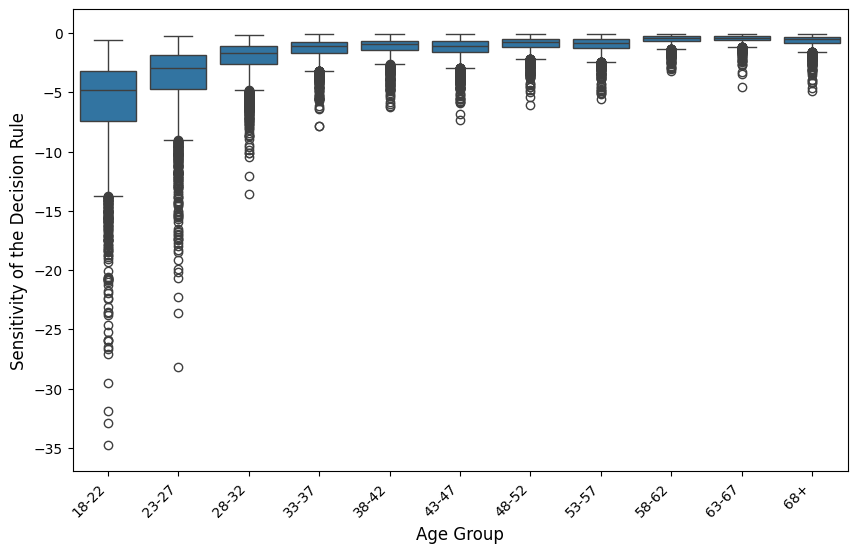}
    \includegraphics[width=0.48\linewidth]{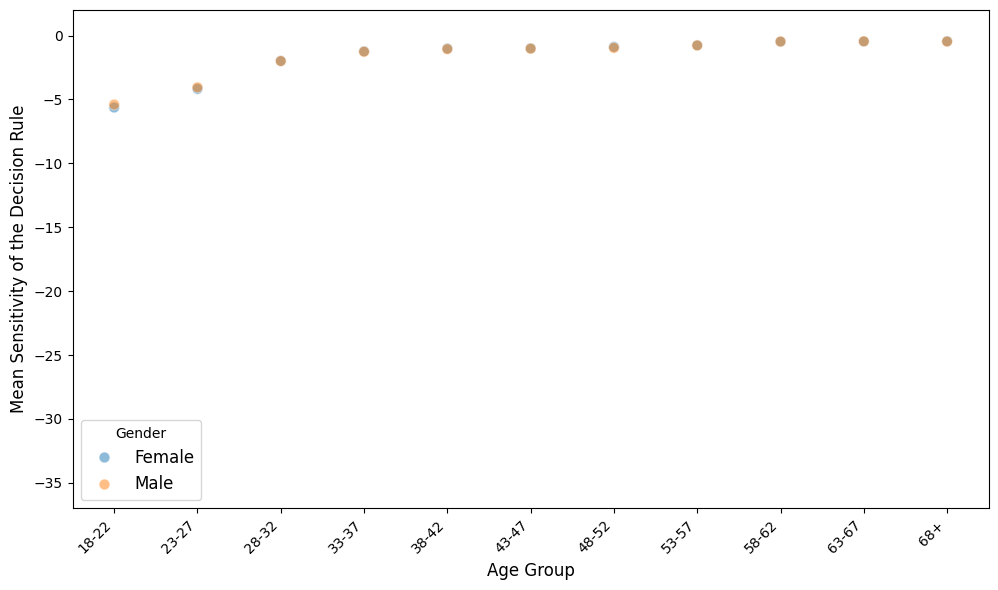}
    \caption{Left: Box plots of the sensitivity of the decision rule for different age groups. Right: average sensitivity of the decision rule for age groups and by gender. Both are under the expected value risk measure.}
    \label{fig:sensitivity_box}
\end{figure}

The left panel of \Cref{fig:sensitivity_box} presents box plots of the sensitivity of the unadjusted conditional expectation decision rule across age groups, while the right panel of \Cref{fig:sensitivity_box} shows the corresponding group-level means disaggregated by gender. Together, these plots illustrate how sensitivity varies with age and help assess potential disparities across demographic subgroups. We observe that younger individuals, particularly those in the 18–27 age range, exhibit markedly higher (i.e., more negative) sensitivity values, suggesting that small perturbations in the protected attribute (gender) can have a larger impact on decisions for this group. As age increases, the sensitivity diminishes in magnitude and stabilizes across age groups, indicating reduced responsiveness of the decision rule to gender-based perturbations. Notably, the group-level mean sensitivities in the right panel of \Cref{fig:sensitivity_box} show close alignment between males and females, highlighting that—on average—the implemented decision rule does not differentially respond to gender within age brackets. These findings reinforce the need for fairness interventions that explicitly target sensitivity, particularly among younger cohorts where the risk of indirect discrimination is more pronounced.

To assess predictive segmentation, we plot Gini curves for each decision rule following the methodology in \cite{goldburd2016generalized}. The Gini index is a standard metric for evaluating the lift of an insurance rating plan—its ability to stratify policyholders from best to worst risks. In this context, lift reflects the effectiveness of the model in assigning actuarially fair premiums, thereby mitigating adverse selection. As a relative measure, the Gini index is typically used to compare the segmentation strength of competing models. To compute it, the dataset is sorted by predicted loss cost (from lowest to highest risk), and the cumulative percentage of exposures and corresponding actual losses are plotted to form the Lorenz curve. The Gini index is then defined as twice the area between this curve and the 45-degree line of equality. A higher Gini index indicates stronger segmentation power.

\begin{figure}[th!]
        \centering
        \includegraphics[width=0.4\linewidth]{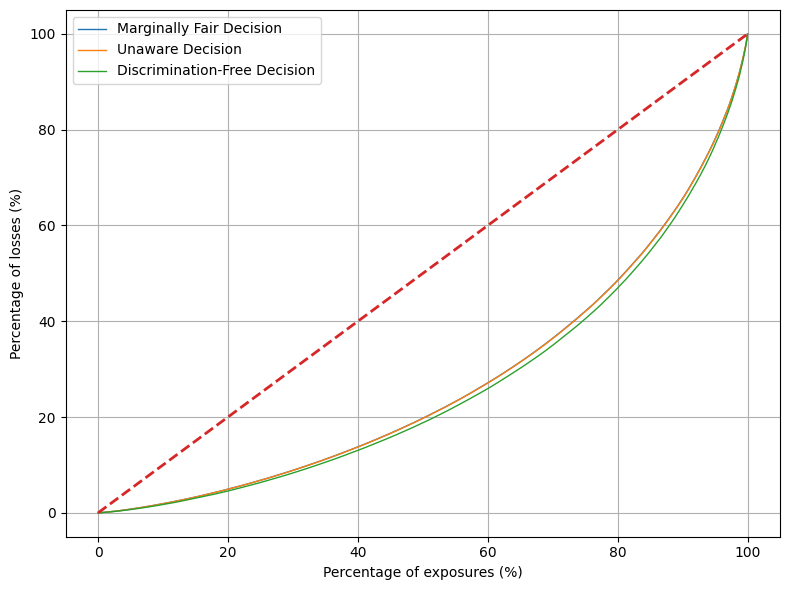}
        \caption{Gini curve comparisons of fair decision strategies under the expected value risk measure.}
        \label{fig:gini}
    \end{figure}

\Cref{fig:gini} presents the Gini curves for the three decision rules. The marginally fair decision rule demonstrates segmentation power on par with both the unaware and discrimination-free rules, indicating that fairness can be enforced without sacrificing risk differentiation—a key requirement in actuarial pricing and underwriting.

\begin{figure}
    \centering
    \includegraphics[width=0.6\textwidth]{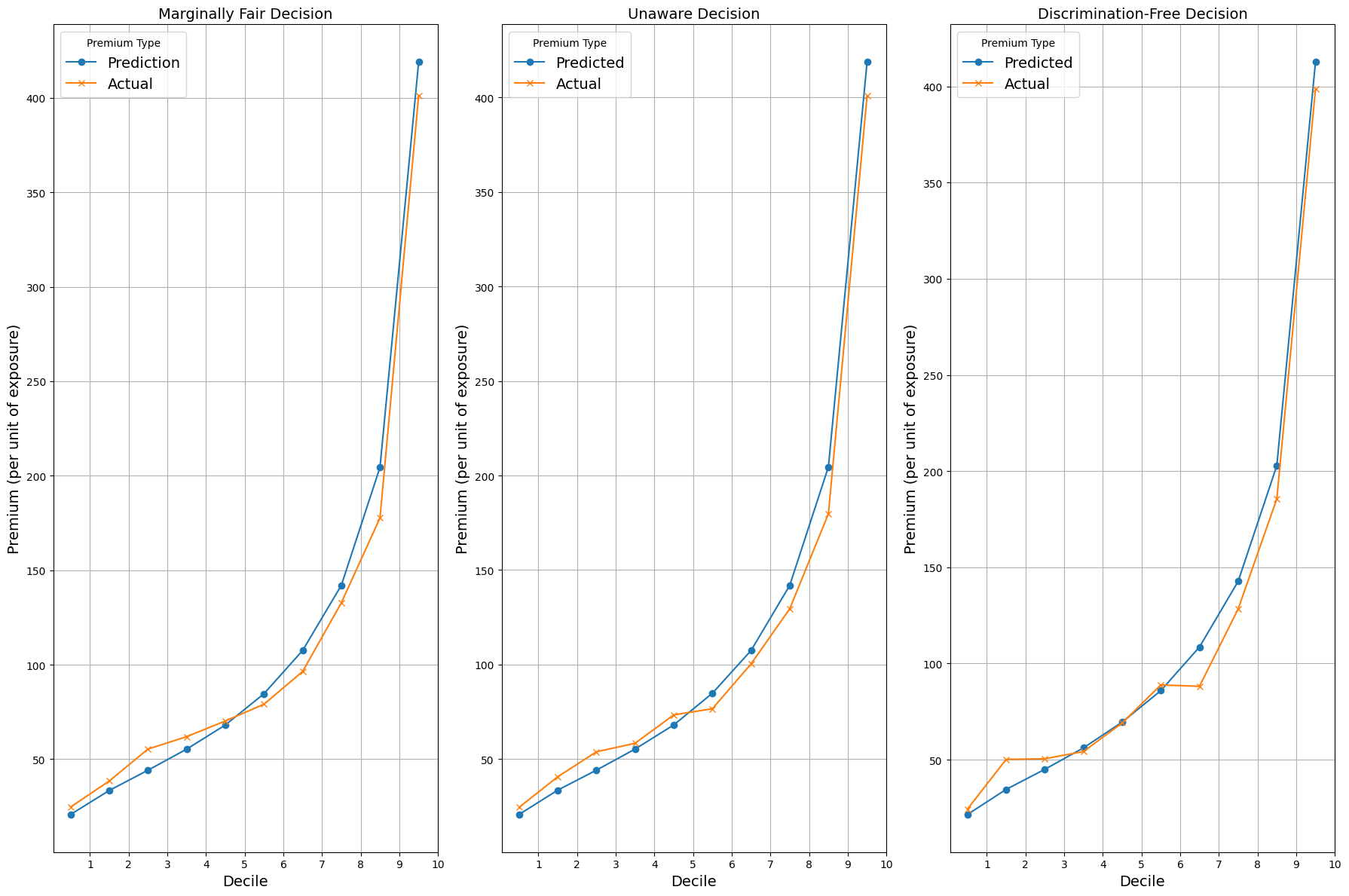}
    \caption{Quantile plot comparing predicted (blue) and observed (orange) losses across decision rules under the expected value risk measure. The x-axis shows exposure-weighted quantiles of predicted claim costs, and the y-axis reports average observed and predicted losses within each bin.  Left panel: marginally fair, middle panel: unaware decision, and right panel: discrimination-free decision. }
    \label{fig:quantile}
\end{figure}

Quantile plots are useful for evaluating predictive models in insurance \cite{goldburd2016generalized}. They assess predictive accuracy by checking how closely the predicted and actual (observed outcomes against each bin) outcomes align within each quantile, where a well-calibrated model has minimal over- or underestimation across all bins. They also evaluate model fit by examining the ratio between the highest and lowest bins, where a larger ratio indicates better segmentation between good and poor risks. Monotonicity is another key aspect, requiring predicted outcomes to increase with higher quantiles, with actual outcomes ideally following a similar trend. To construct these plots, the dataset is sorted by predicted loss cost, divided into equal-exposure quantiles (e.g., deciles), and the average predicted and observed pure premiums are computed within each bin. \Cref{fig:quantile} presents quantile plots comparing predicted and actual outcomes across decision rules. The marginally fair rule achieves comparable calibration and segmentation power, suggesting that fairness can be achieved without sacrificing predictive accuracy or model fit.

These empirical results mirror the findings from our numerical study: marginal fairness adjustments can be implemented feasibly in practice, yielding marginally fair decisions while maintaining strong predictive and segmentation performance. 
Thus, marginal fairness offers a practical approach to achieving fairness in real-world applications.



\subsection{Expected Shortfall}
In this subsection we discuss the modeling and results based on the ES at the 90\% level. In practice, this could correspond to the additional risk loading that needs to be charged to policyholders to account for tail risk.

The implementation of the marginally fair ES follows a similar structure to that of the expected value risk measure discussed in the prior subsections. The key differences lies in estimating ES and its sensitivity. To estimate ES, \( \rho_\gamma(Y \mid X) \), we first fit a quantile regression model to estimate the Value-at-Risk (VaR) at the 90\% level using \( X \). Recall that VaR at level $u \in (0,1)$ is equal to the quantile function evaluated at $u$, i.e. for $Y \in \L$, it holds $\VaR_\alpha(Y):= F_Y^{-1}(u)$. Next, we fit a GLM with Tweedie loss, conditional on the values of \( X \) that exceed the estimated VaR. To estimate the sensitivity of the ES risk measure, we apply \Cref{thm:sens-discrete}. Specifically, we use a NN with binary cross-entropy loss to predict an indicator variable defined as the product of the gender variable and the weight function, using only \( X \) as input.
Model specifications for each component are detailed in \Cref{app:model}.

\begin{figure}
    \centering
    \includegraphics[width=0.6\linewidth]{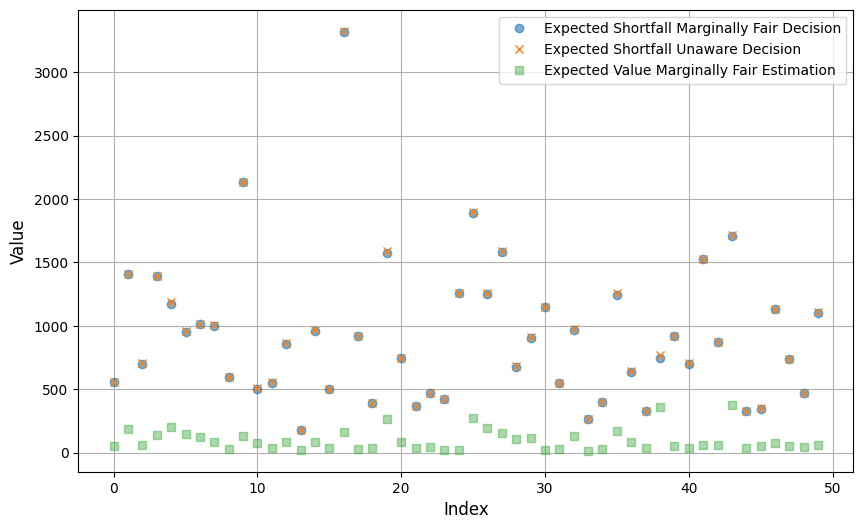}
    \caption{Comparison of fair decision strategies for 50 randomly selected policyholders under ES. Blue dots correspond to marginally fair decisions under ES, orange crosses to unaware decision under ES, and the green squares to the marginally fair decision under the expected value risk measure. }
    \label{fig:50points_ES}
\end{figure}

Figure~\ref{fig:50points_ES} compares three decision rules: the marginally fair decision under the ES, the unaware decision under ES, and the marginally fair decision under the expected value risk measure. The first two decision rules, the marginally fair and unaware ES decisions,  yield similar values for most policyholders, although visible discrepancies occur in certain cases. In contrast, the marginally fair decisions based on the expected value risk measure are, as expected, substantially lower across the board. It is also worth noting that, because ES values are generally much larger, the vertical scale of this plot is significantly greater than its expected value counterpart, making visual differences between the two ES strategies more difficult to discern.

Table~\ref{tbl:summaryDecisionES} shows that the marginally fair decisions have slightly lower values than the unaware decisions at different quantiles, suggesting that the fairness adjustments are modest at the aggregate level. However, Table~\ref{tbl:summaryPU-PMFES} reveals that the individual-level differences \( P_U - P_{MF} \) can be substantial, with a maximum exceeding 200, even though the median is only 3.28. This pattern indicates that marginal fairness can correct decisions for individuals who would otherwise be most affected by indirect discrimination, while maintaining overall alignment with the original risk-based pricing structure.

\begin{table}[ht!]
\centering
\scalebox{0.9}{
\begin{tabular}{lccccc}
\toprule
\textbf{Decision Rule}  & \textbf{Min}  & \textbf{25\% Quantile} & \textbf{50\% Quantile} & \textbf{75\% Quantile} & \textbf{Max} \\
\midrule 
Marginally Fair, $P_{MF}$      & 0.04  & 567.76 & 868.90 & 1258.28 & 5728.73 \\
Unaware, $P_U$                  & 0.94  & 570.18 & 873.14 & 1265.82 & 5738.26 \\
\bottomrule
\end{tabular}}
\caption{Summary statistics of decisions under different fairness criteria under ES}
\label{tbl:summaryDecisionES}
\end{table}

\begin{table}[H]
\centering
\scalebox{0.9}{
\begin{tabular}{lccccc}
\toprule
\textbf{Statistic} & \textbf{Min} & \textbf{25\% Quantile} & \textbf{50\% Quantile} & \textbf{75\% Quantile} & \textbf{Max} \\
\midrule
$P_U - P_{MF}$  & 0.06 & 1.79 & 3.28 & 6.37 & 204.82 \\
\bottomrule
\end{tabular}}
\caption{Summary statistics of the difference between unaware and marginally fair decisions, $P_U - P_{MF}$, under ES}
\label{tbl:summaryPU-PMFES}
\end{table}

\begin{figure}
    \centering
    \includegraphics[width=0.44\linewidth]{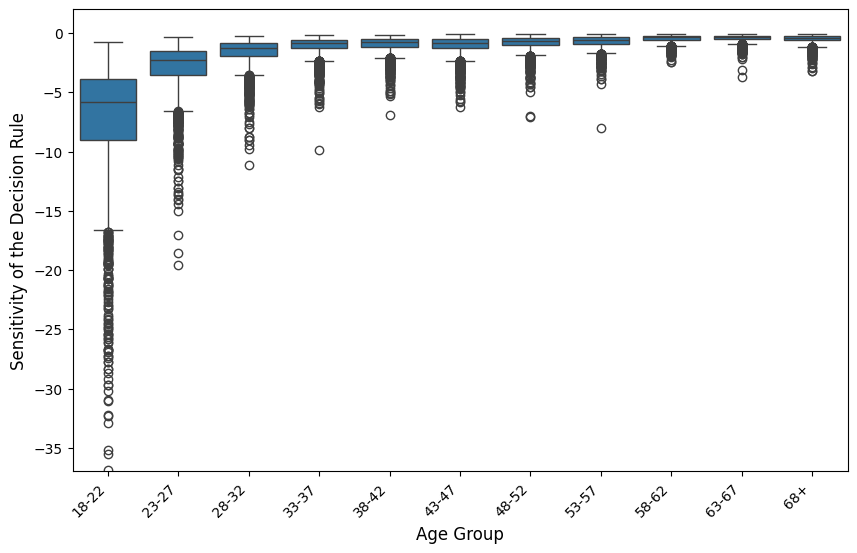}
    \includegraphics[width=0.48\linewidth]{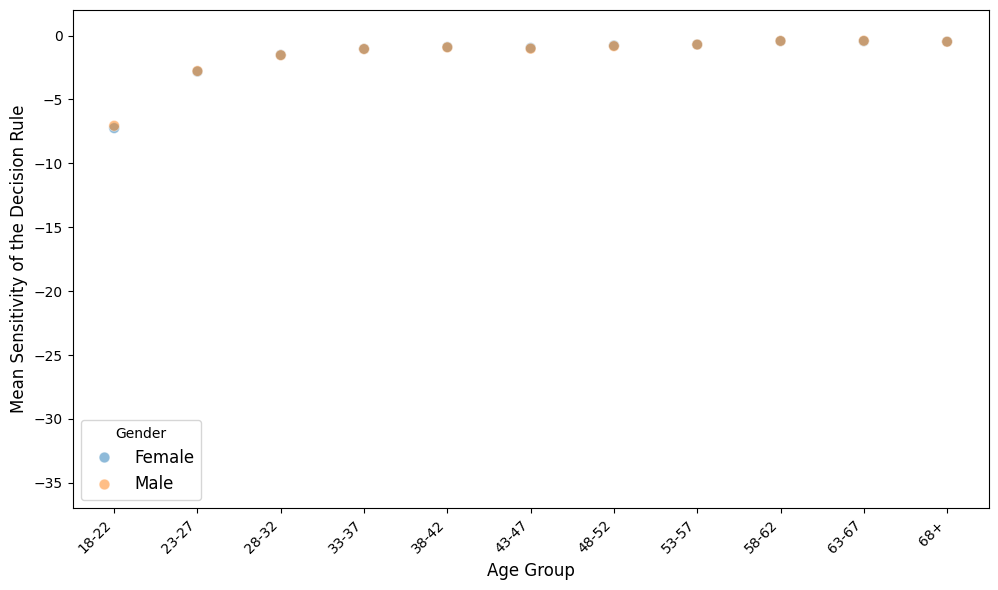}
    \caption{Left: Box plots of the sensitivity of the decision rule for different age groups. Right: average sensitivity of the decision rule for age groups and by gender. Both under ES. }
    \label{fig:sensitivity_box_ES}
\end{figure}

The sensitivity of the ES decision rule across different age groups exhibits similar patterns to that of the expected value decision rule. However, the magnitude of sensitivity is generally larger under the ES risk measure, particularly for younger age groups. This suggests that decisions based on tail risk measures are more sensitive to perturbations in protected attributes among younger individuals. These patterns are illustrated in Figure~\ref{fig:sensitivity_box_ES}.

\section{Conclusions} \label{sec:conclusion}
This paper introduces marginal fairness as a new individual fairness criterion for decision-making under risk, ensuring that final outcomes are locally insensitive to protected attributes. By modeling decision-making as a two-step process—first predicting the outcome and then applying a generalized distortion risk measure—we provide a flexible and realistic framework that aligns with real-world practices in insurance, finance, and other high-stakes fields. Our theoretical development shows that fairness should be enforced at the decision stage and we propose methods to achieve marginal fairness across continuous, bounded, discrete, and multivariate protected variables. Furthermore, we extend the framework to incorporate cascade sensitivity, accounting for statistical dependencies among covariates.

Through a numerical study and an empirical case study on auto insurance data, we demonstrate that marginal fairness can be achieved with minimal sacrifice in predictive accuracy and segmentation efficiency. Our results suggest that it is possible to design decision rules that are both fair and effective, offering a practical solution to regulatory and ethical challenges in algorithmic decision-making.

While the framework developed in this paper offers an operationally feasible approach to fair decision-making, several limitations merit further investigation. First, our methodology assumes access to accurate predictive models and reliable estimates of sensitivity measures, which may be challenging in settings with limited data or complex dependence structures. Second, while cascade sensitivity captures statistical dependencies, it does not account for potential causal relationships, which could provide deeper insights into fairness interventions. Finally, future research directions also include developing testing tools for marginal fairness evaluation and integrating welfare-based objectives into fair decision rules.

\section*{Acknowledgement}
FH gratefully acknowledges support from the Australian Research Council with funding reference number DP250104816. SP gratefully acknowledges support from the Natural Sciences and Engineering Research Council of Canada with funding reference numbers DGECR-2020-00333 and RGPIN-2020-04289. We also thank our research assistant, Md Mushahidul Islam Shamim, for his valuable help with the data analysis.

\begin{appendix}
\section{Additional details on Examples}\label{app:ex}

\subsection*{\Cref{ex:Y-D-dependent-zero-sens}}\label{proof-Ex}

For completeness we derive the cdf and quantile function of $Y$. Using the definition of $Y$, and independence of $(D, X_1, X_2)$ in the second equation, we obtain
    \begin{align*}
        F_Y(y) 
        &= 
        \P(Y \le y | X_1 = 0)(1-p) + \P(Y \le y | X_1 = 1)\,p
        \\
        &= 
        \P(D \le y)(1-p) + \P(X_2 \le y )\,p
        \\
        &=
        \Id_{\{y \le C\}} \,(1-p)F_D(y) + \Id_{\{y > C\}} \,\big(1-p + pF_{X_2}(y)\big)\,.
    \end{align*}
Next, the quantile function of $Y$ is
\begin{align*}
    \Finv_Y(u)
    &= 
    \inf\{ y \in \R \, |\, F_Y(y) \ge u\}
    \\
    &= 
    \inf\{ y \in \R \, |\, \Id_{\{y \le C\}} \,(1-p)F_D(y) + \Id_{\{y > C\}} \,\big(1-p + pF_{X_2}(y)\big) \ge u\}
    \\
    &= \begin{cases}
        \Finv_D\big(\frac{u}{1-p}\big) \qquad  & u \le 1-p
        \\[1em]
        \Finv_{X_2}\big(\frac{u - 1 + p}{p}\big) & u > 1-p\,.
    \end{cases}
\end{align*}
To see the last equality, note that if $(1-p)F_D(C) \ge u$ (which is equivalent to $1-p \ge u$), then the infimum simplifies to $\inf\{ y \in \R \, |\, (1-p)F_D(y) \ge u\} = \Finv_D(\frac{u}{1-p})$. The case when $(1-p)F_D(C) > u$ follows similarly.

\subsection*{\Cref{mortgage-cascade}}
The distribution of $F_{X|D}(x|t)$ is given by 
\begin{align*}
    F_{X|D}(x\, |\, t)
    &= 
    \P(X \le x | D = t)
    \\
    &=
    \P(X \le x | D = 0)\Id_{\{t = 0\}} + \P(X \le x | D = 1)\Id_{\{t = 1\}}
    \\
    &=
    \Phi\left(\tfrac{\log(x) - \mu}{\sigma}\right)\Id_{\{t = 0\}}
    + \Phi\left(\tfrac{\log(x) - 2\mu}{\sigma}\right)\Id_{\{t = 1\}}\,.
\end{align*}
Moreover, the conditional quantile function of $F_{X|D}(x\, |\, t)$ (the inverse in $x$ with $t$ fixed) is
\begin{align*}
    \Finv_{X|D}(u\, |\, t)
    &=
    \exp\left\{ \Phi^{-1}(u)\sigma  + \mu \right\}\Id_{\{t = 0\}}
    +
    \exp\left\{ \Phi^{-1}(u)\sigma  + 2 \mu \right\}\Id_{\{t = 1\}}\,.
\end{align*}
Next, the perturbation on $X$ induced by $D_\delta$ is 
\begin{align*}
    X_\delta
    :&= 
    \Finv_{X|D}(V\, |\, t)\big|_{t = D_\delta}
    =
    \exp\left\{ \Phi^{-1}(V)\sigma  + \mu \right\}\Id_{\{D_\delta = 0\}}
    +
    \exp\left\{ \Phi^{-1}(V)\sigma  + 2 \mu \right\}\Id_{\{D_\delta = 1\}}\,,
\end{align*}
where $V\sim U(0,1)$ and the cdf of $X_\delta$ is 
\begin{equation*}
    F_{X_\delta}(x)
    =
    \Phi\left(\tfrac{\log(x) - \mu}{\sigma}\right)(1-p_\delta)
    + \Phi\left(\tfrac{\log(x) - 2\mu}{\sigma}\right)p_\delta,.
\end{equation*}

\section{Auxiliary results}\label{app:aux-results}

The next results are generalizations of lemma 1 in \cite{Pesenti2024EJOR-non-diff}.

\begin{lemma}\label{lemma:delta-function-general}
Let $\kappa_\delta(z)$, $\delta \ge 0, z \in\R$ be a function that is differentiable in both $z$ and $\delta$ and satisfies $\kappa_0(z) = z$ for all $z \in \R$. For fixed $p \in \R$ and $\delta > 0$, define the family of functions
    \begin{equation}
    \label{eq:h-delta}
        h_\delta(z) := \frac{1}{\delta} \, \big(\Id_{\{\kappa_\delta(z) \le p\}} - \Id_{\{z \le p\}} \big)\,, \quad z\in \R\,.
    \end{equation}
Then, for any measurable function $H \colon \R^k \to \R$ and rv $Z$ and random vector $\W$, such that $  \E[H(\W) \, |\, \X ]< +\infty$, it holds
 \begin{equation*}
     \lim_{\delta \downarrow 0} \E[h_\delta(Z) H(\W) \, |\, \X ]
     =
     \E[\frac{\partial}{\partial_\delta} \kappa_\delta^{-1}(Z)\big|_{\delta = 0} H(\W) \Id_{\{Z = p\}}\, |\, \X ]\,,
 \end{equation*}
 where $\kappa_\delta^{-1}(z)$ denotes the inverse of $\kappa_\delta(z)$ with respect to $z$. 
\end{lemma}
\begin{proof}
Let $\xi$ be an infinitely often differentiable function. Using the change of variable $y = \kappa_\delta(z)$   
\begin{align*}
        \int_\R\xi(z) h_\delta(z) \diff z
        &= 
        \frac1\delta \int_\R\xi(z) \big(\Id_{\{\kappa_\delta(z)\le p \}} - \Id_{\{z \le p\}} \big) \diff z
        \\
        &= 
        \frac1\delta \int_\R \frac{\xi(z)}{\frac{\partial}{\partial z'} \kappa_\delta(z')} \Big|_{z' =z= \kappa_\delta^{-1}(y)} \Id_{\{y \le p \}}  \diff y
        -
        \frac1\delta \int_{-\infty}^{p}\xi(z) \diff z\,.
    \end{align*}
Letting $\Xi$ be the primitive of $\xi$ vanishing at $-\infty$, we obtain
\begin{align*}
        \int_\R\xi(z) h_\delta(z) \diff z
        &= 
        \frac1\delta \int_{-\infty}^{p}
        \frac{d}{dz} \Xi \big(\kappa_\delta^{-1}(z)\big) \diff z
        -
        \frac1\delta\, \Xi(p)
        = 
        \frac1\delta\, \big( \Xi(\kappa_\delta^{-1}(p)
         -
        \Xi(p)\big)\,.
\end{align*}
Taking the limit (note that $\kappa_0(x) = x$)
\begin{equation*}
    \lim_{\delta \to 0} \int_\R\xi(z) h_\delta(z) \diff z
    =
    \xi(p) \frac{\partial}{\partial \delta}\kappa_\delta^{-1}(p)\Big|_{\delta = 0}\,.
\end{equation*}

For the second statement, we have
\begin{align*}
    \lim_{\delta \downarrow 0} \E[h_\delta(Z) H(\W) \, |\, \X ]
     =
     \frac{\partial}{\partial \delta}\kappa_\delta^{-1}(p)\big|_{\delta = 0}\; \E[ H(\W) \Id_{\{Z = p\}}\, |\, \X ]\,.
\end{align*}
\end{proof}

\begin{lemma}\label{lemma:delta-function-very-general}
Let $\kappa_\delta(z)$, $\delta \ge 0, z \in\R$ be a function that is differentiable in both $z$ and $\delta$, invertible in $z$, and satisfies $\kappa_0(z) = z$ and $\kappa_0^{-1}(z) = z$ for all $z \in \R$. Further let $\ell\colon \R^{m+1} \to \R$ be differentiable and invertible in its first component. Then, for fixed $p \in \R$ and $\delta > 0$, define the family of functions
    \begin{equation*}
        h_\delta(z;\v, p) := \frac{1}{\delta} \, \big(\Id_{\{\ell(\kappa_\delta(z); \v ) \le p\}} - \Id_{\{\ell(z; \v) \le p\}} \big)\,, \quad z\in \R, \v \in \R^m\,.
    \end{equation*}
Then, for any text function $\xi\colon \R\to \R$, it holds that 
 \begin{equation*}
    \lim_{\delta \to 0} \int_\R\xi(z) h_\delta(z;\v, p) \diff z
    =
    \xi\big(\ell^{-1}(p;\v)\big) \tfrac{\partial}{\partial \delta}\kappa_\delta^{-1}\big(\ell^{-1}(p;\v)\big)\Big|_{\delta = 0}\,.
 \end{equation*}
\end{lemma}
\begin{proof}
Fix $\v \in \R^m$ and let $\xi$ be an infinitely often differentiable function. Using in the first integral the change of variable $y = \ell(\kappa_\delta(z); \v)$, and in the second integral the change of variable $y' = \ell(z; \v)$, where we omit the dependence on $\v$. Then we obtain
\begin{align*}
        \int_\R\xi(z) h_\delta(z; \v, p) \diff z
        &= 
        \frac1\delta \int_\R\xi(z) \Id_{\{\ell(\kappa_\delta(z); \v)\le p \}} \diff z
        -
        \frac1\delta \int_\R\xi(z) \Id_{\{\ell(z; \v) \le p\}} \diff z
        \\
        &= 
        \frac1\delta \int_{-\infty}^p \frac{\xi(z)}{\frac{\partial}{\partial z'}\big( \ell(\kappa_\delta(z'), \v)\big)} \Big|_{z' = z= \kappa_\delta^{-1}(\ell^{-1}(y; \v))} \diff y
        \\
        & \quad
        -
        \frac1\delta \int_{-\infty}^{p}\frac{\xi(z)}{\frac{\partial}{\partial z}\ell(z; \v)}\Big|_{z = \ell^{-1}(y'; \v)} \diff y'\,.
    \end{align*}
Letting $\Xi$ be the primitive of $\xi$ vanishing at $-\infty$, it holds 
\begin{align*}
        \int_\R\xi(z) h_\delta(z; \v, p) \diff z
        &= 
        \frac1\delta \int_{-\infty}^{p}
        \frac{d}{dz} \Xi \big(\kappa_\delta^{-1}(\ell^{-1}(z; \v))\big) \diff z
        -
        \frac1\delta \int_{-\infty}^{p}
        \frac{d}{dz} \Xi \big(\ell^{-1}(z; \v)\big) \diff z
        \\
        &= 
        \frac1\delta\, \Big[ \, \Xi\big(\kappa_\delta^{-1}\big(\ell^{-1}(p; \v)\big)\big)
         -
        \Xi\big(\ell^{-1}(p;\v)\big)\Big]\,.
\end{align*}
Taking the limit (recall that $\kappa^{-1}_0(x) = x$)
\begin{equation}\label{eq:delta-limit-general}
    \lim_{\delta \to 0} \int_\R\xi(z) h_\delta(z;\v, p) \diff z
    =
    \xi\big(\ell^{-1}(p;\v)\big) \tfrac{\partial}{\partial \delta}\kappa_\delta^{-1}\big(\ell^{-1}(p;\v)\big)\Big|_{\delta = 0}\,.
\end{equation}
\end{proof}

\begin{lemma}\label{lemma:derivative-kappa}
Define the function $\kappa_\delta(u): = \Phi(\Phi^{-1}(u)(1 + \delta)\big)$, for $\delta \ge 0$ and $u \in (0,1)$. Then, 
 \begin{equation*}
     \frac{\partial}{\partial \delta} \kappa_\delta^{-1}(u)\Big|_{\delta = 0}
     =
     - \Phi^{-1}(p) \,\phi\big(\Phi^{-1}(p)\big)\,.
 \end{equation*}
\end{lemma}
\begin{proof}
Note that $\kappa_\delta^{-1}(u) = \Phi\big(\tfrac{1}{1 + \delta}\, \Phi^{-1}(u)\big)$. We calculate, by making the change of variable $y:= \Phi^{-1}(u)$,
\begin{align*}
\lim_{\delta \to 0} \tfrac{1}{\delta} \big(\kappa_\delta^{-1}(u)  - \kappa_0^{-1}(u) \big)
    &=
    \lim_{\delta \to 0} \tfrac1\delta \Big\{\Phi\big(\tfrac{1}{1 + \delta}\, \Phi^{-1}(u)\big) - \Phi \big(\Phi^{-1}(u)\big)\Big\}
    \\
&= 
\lim_{\delta \to 0} \tfrac1\delta \Big\{\Phi\big(\tfrac{1}{1 + \delta}\, y\big) - \Phi (y)\Big\}
\\
&=
\lim_{\delta \to 0} \tfrac1\delta \Big\{\Phi\Big( y (1 - \delta) + o(\delta) \Big) - \Phi (y)\Big\}
\\
&=
\lim_{\delta \to 0} \tfrac1\delta \big\{\Phi(y)  - \delta y\, \phi(y)  - \Phi (y) +  o(\delta)\big\}
\\
&= 
- y \phi(y)
\\
&=
- \Phi^{-1}(u) \,\phi\big(\Phi^{-1}(u)\big)\,,
\end{align*}
where we used in the third equality the Taylor approximation of $\frac{1}{1 + \delta}$ and in the forth the Taylor approximation of $\Phi(\cdot)$ around $y$. 
\end{proof}

\section{Proofs}\label{app:proofs}

\begin{proof}[Proof of \Cref{prop:marginal-sensitivity}]\label{proof-sensitivity-rep}
This proof follows by a generalization of Prop. 11 in \cite{pesenti2024risk}, which only holds for linear prediction function $\mfg$. We also refer to \cite{Tsanakas2016RA} for an alternative proof for the sensitivity measure of unconditional distortion risk measures. 

To simplify notation, we write the distorted output $Y_\delta:=\mfg\big( \D_{i,\delta} , \X\big) $, where $\D_{i,\delta}:= (D_1, \ldots, D_{i-1}, D_i(1 + \delta), D_{i+1}, \ldots, D_m)$, the conditional cdfs $F(y):= \P(Y \le y |\X = \x, \D_{-i} = \d_{-i})$, and $F(y, \delta):= \P(Y_\delta \le y |\X = \x, \D_{-i} = \d_{-i})$. Then we have that 
\begin{equation*}
\rho_\gamma\big(Y_\delta\,|\,\X = \x\big) = \E\Big[\E[F^{-1}(U, \delta) \gamma(U) | \X = \x, \D_{-i} = \d_{-i}]\, \Big| \, \X = x\Big]\,    
\end{equation*}
for a, conditional on $(\X = \x, \D_{-i} = d_{-i})$, uniform rv $U$. Moreover, using the mean value theorem together with Lebesgue dominated convergence we interchange expectation and limit to 
\begin{align}
\lim_{\delta \to 0}\frac{
    \rho_\gamma\big(Y_\delta\,|\,\X = \x\big) - \rho_\gamma\big( Y\,|\,\X=\x\big)}{\delta}
    \nonumber
    \\
    &\hspace*{-6em}=
    \E\Big[\E[\partial_\delta F^{-1}(U, \delta) \big|_{\delta = 0}\gamma(U) | \X = \x, \D_{-i} = \d_{-i}]\, \Big| \, \X = \x\Big]
    \nonumber
    \\
    &\hspace*{-6em}=
    \E\Big[\E[\partial_\delta F^{-1}(F(Y), \delta) \big|_{\delta = 0}\gamma(F(Y)) | \X = \x, \D_{-i} = \d_{-i}]\, \Big| \, \X = \x\Big]
    \nonumber
    \\
    &\hspace*{-6em}=
    \E\Big[\int_\R \partial_\delta F^{-1}(F(y), \delta) \big|_{\delta = 0}\gamma(F(y)) f(y)\diff y \Big| \, \X = \x\Big]
    \,,
    \label{eq:derivative-rho-proof}
\end{align}
where $\partial_\delta := \tfrac{\partial}{\partial_\delta} $ denotes the partial derivative with respect to $\delta$, $f$ the density of $Y_\delta$ given $(\X, \D_{-i})$, and where we used that $\P$-a.s. $U = F(Y)$, given $(\X, \D_{-i})$.  
By taking derivative of the equation $F(F^{-1}(U, \delta), \delta) = u$, we obtain for all $u \in (0,1)$ that
\begin{equation*}
    \tfrac{\partial}{\partial_\delta} F^{-1}(u, \delta)\Big|_{\delta = 0}
    =
    -\frac{\partial_\delta F(y, \delta)}{f(y)}\Big|_{y = F^{-1}(u)}
\end{equation*}
Inserting the representation of $\partial_\delta F(y, \delta)$ into \eqref{eq:derivative-rho-proof}, the integral becomes
\begin{align*}
    \int_\R \partial_\delta F^{-1}(F(y), \delta) \big|_{\delta = 0} \gamma(F(y)) f(y)\diff y
    &=
     \int_\R -\frac{\partial_\delta F(y, \delta)}{f(y)}\gamma(F(y)) f(y)\diff y
     \\
     &=
     \int_\R -\partial_\delta F(y, \delta)\gamma(F(y))\diff y
     \,. 
\end{align*}
Next, we applied \Cref{lemma:delta-function-very-general} with $p = y$, $\ell = \mfg$, and $\kappa_\delta(x) = x(1 + \delta)$ to 
\begin{align*}
    \partial_\delta F(y, \delta)
    &=
    \lim_{\delta \to 0}\, \tfrac1\delta\, \E[\Id_{\{\mfg( \D_{i,\delta} , \X)  \le y\}} - \Id_{\{\mfg( \D , \X)  \le y\}} |\X = \x, \D_{-i} = \d_{-i}]
    \\
    &= \tfrac{\partial}{\partial \delta } \kappa_\delta^{-1}\big(\mfg^{-1}(y, \d_{-i}, \x)\big)\big|_{\delta = 0}\; f_{D_i | \X, \D_{-i}}\big(\mfg^{-1}(y, \d_{-i}, \x)\big),
\end{align*}
where $\mfg^{-1}$ denotes the inverse in its $i$-th component, and we use the notation $\mfg^{-1}(y, \d_{-i}, \x): = \mfg^{-1}(d_1, \ldots, d_{i-1}, y, d_{i+1}, \ldots, d_m, \x)$.  
Noting that $\partial_\delta \kappa^{-1}_\delta(x)|_{\delta = 0} = - x$, and then making a change of variable $t = \mfg^{-1}(y, \d_{-i}, \x)$, which implies that $\partial_i \mfg(t, \d_{-i}, \x) \diff t= \diff y$
\begin{align*}
    \int_\R \partial_\delta F^{-1}(F(y), \delta) \big|_{\delta = 0}& \gamma(F(y)) f(y)\diff y
    \\
    &=
    \int_\R \mfg^{-1}(y, \d_{-i}, \x)\; f_{D_i | \X, \D_{-i}}\big(\mfg^{-1}(y, \d_{-i}, \x)\big)\gamma(F(y))\diff y
    \\
    &=
    \int_\R t\;\partial_i \mfg(t, \d_{-i}, \x) \;  \gamma\big(F(\mfg(t, \d_{-i}, \x))\big)\, f_{D_i | \X, \D_{-i}}(t)\, \diff t
    \\
    &= 
    \E\big[D_i\;\partial_i \mfg(\D, \X) \; \gamma\big(F(\mfg(\D, \X))\big)\, |\, \X = \x, \D_{-i} = \d_{-i}\big]
    \\
    &= 
    \E\big[D_i\;\partial_i \mfg(\D, \X) \; \gamma(F(Y))\, |\, \X = \x, \D_{-i} = \d_{-i}\big].
\end{align*}
Collecting, we obtain 
\begin{align*}
    \lim_{\delta \to 0}\frac{
    \rho_\gamma\big(Y_\delta\,|\,\X = \x\big) - \rho_\gamma\big( Y\,|\,\X=\x\big)}{\delta}
    =
     \E\big[D_i\;\partial_i \mfg(\D, \X) \, \gamma(F(Y))\, |\, \X = \x\, \big]\,.
\end{align*}
\end{proof}

\begin{proof}[Proof of \Cref{thm:individual}]
For simplicity, we omit the superscripts of $\ell^{\t, \x}\in \Gamma^{\t, \x}$ and simply write $\ell$. Next for any uniform rv $U\sim U(0,1)$, the objective function in \eqref{opt:main} is
\begin{align*}
    \int_0^1 \big(\gamma(u) - \ell(u)\big)^2\, du
    &= \E\big[\, (\gamma(U) - \ell(U)\big)^2 \, \big]
    \\
     &= \E\big[\, \E\big[\big(\gamma(U_{Y|\X}) - \ell(U_{Y|\X})\big)^2 \,| \X\big]\, \big]
     \\
     &= 
     \int_{\R^n}\int_{\R^{m}} \big(\gamma(U_{\mfg(\t, \x)|\X}) - \ell(U_{\mfg(\t, \x)|\X})\big)^2\,\diff F_{\D|\X}(\t|\x)\  \,  \diff F_\X(\x)\,,
\end{align*}
where the second equation follows by choosing the uniform rv to be $U_{Y|\X}$ and where for $(\x, \t) \in \R^{n+m}$, we define $U_{\mfg(\t, \x)|\X}:= F_{\mfg(\D, \X)|\X}\big(\mfg(\t, \x) | \x\big)$.

Using \Cref{prop:marginal-sensitivity}, we define the Lagrangian, pointwise in $\x$, of optimisation problem \eqref{opt:main} with Lagrange multipliers $\eta(\x) \in \R $ by
\begin{align*}
    L(\ell, \x)
    :&=
    \int_{\R^{m}}\Big\{ \Big(\ell(U_{\mfg(\t, \x)|\X}) - \gamma(U_{\mfg(\t, \x)|\X})\Big)^2
    + 2\eta(\x)\,  t_i \, \partial_{i} \mfg(\t, \x) \ell(U_{\mfg(\t, \x)|\X}) 
      \Big\}\,\diff F_{\D|\X}(\t|\x)\  \, 
    \\
    &= 
    \int_{\R^{m}}\Big\{ \Big( \ell(U_{\mfg(\t, \x)|\X})- \Big[\gamma(U_{\mfg(\t, \x)|\X}) - \eta(\x)\,  t_i \, \partial_{i} \mfg(\t, \x) 
   \Big]\Big)^2 
    \\
    & \qquad \qquad
    - \Big[\gamma(U_{\mfg(\t, \x)|\X}) - \eta(\x)\,  t_i \, \partial_{i} \mfg(\t, \x) \Big]^2
    +
    \gamma(U_{\mfg(\t, \x)})^2
    \Big\}\,\diff F_{\D|\X}(\t|\x)\, .
\end{align*}
Using pointwise optimisation, a solution has to satisfy for all $(\t, \x) \in \supp(\D, \X)$ 
\begin{equation} \label{eq:h-opt}
    \ell^*_{\eta(\x)}(U_{\mfg(\t, \x)|\X})=
    \gamma(U_{\mfg(\t, \x)|\X}) - \eta(\x)\,  t_i \, \partial_{i} \mfg(\t, \x)\,.
\end{equation}
Next, we calculate the Lagrange multiplier. For this we enforce for each $\x$ the constraint
\begin{subequations}\label{eqs:Lagrange-multiplier}
    \begin{align}
    0 &=\partial_{D_i}\;  \rho_{\ell^*_{\eta(\x)}}(\, Y\,|\,\X = \x\,)
        \\
        &=
        \E\big[\,D_i \,\partial_{i} \mfg(\D, \X) \ell^*_{\eta(\x)}\big(U_{Y|\X}\big)  \,|\, \X = \x\,\big]
        \\
        &=
        \partial_{D_i}\;  \rho_{\gamma}(\, Y\,|\,\X = \x\,)
        -
         \eta(\x)\,\E\big[\,\big(D_i \,\partial_{i} \mfg(\D, \X) \big)^2  \,|\, \X = \x\,\big]\,.
    \end{align}
\end{subequations}

Thus, the optimal Lagrange multiplier is
\begin{equation*}
    \eta^*(\x)  =  \frac{\partial_{D_i}\;  \rho_{\gamma}(\, Y\,|\,\X = \x\,)}{\E\big[\,\big(D_i \,\partial_{i} \mfg(\D, \X) \big)^2  \,|\, \X = \x\,\big]}\,.
\end{equation*}
\Cref{eq:optimal-h} holds by replacing $\eta(\x)$ in \eqref{eq:h-opt} by the optimal Lagrange multiplier $\eta^*(\x)$, yielding $\ell^*_{\eta^*(\x)}$. By \Cref{asm-integrability}, $\ell^*_{\eta^*(\x)}$ is square integrable in $(\t, \x)$ and hence belongs to $\Gamma^{\t,\x}$.

The marginally fair premium follows by explicitly calculating
\begin{align*}
    \rho_{\ell^*_{\eta^*(\x)}}(Y \,|\,\X) 
       &=
       \E\big[Y \ell^*_{\eta^*(\x)} (U_{Y|\X})\, |\X\big]
       \\
    &=
    \rho_\gamma(Y \,|\,\X) -
    \frac{ \partial_{D_i}\;  \rho_\gamma(\, Y\,|\,\X\,)}{\E\big[\,\big(D_i \partial_{i} \mfg(\D, \X) \big)^2 \big|\,\X\big]}\; 
    \E[Y \, D_i \, \partial_{i} \mfg(\D, \X)\, |\,\X ]    \,.
\end{align*}
Uniqueness follows since the optimization problem is strictly convex in $\ell$ and the constraint is linear in $\ell$. Existence follows by existence of the Lagrange multiplier. Setting the notation $\gamma^*:= \ell^*_{\eta^*(\x)}$ concludes the proof.
\end{proof}

\begin{proof}[Proof of \Cref{prop:mutli-marginal}]
    We proceed similarly to the proof of \Cref{thm:individual} in that we consider the
Lagrangian, pointwise in $\x$, of the optimization problem \eqref{opt:main} with constraints $ \partial_{D_i}\;  \rho_\ell(\, Y\,|\,\X\,)= 0$, $i = 1\ldots, m$. Indeed the Lagrangian is
\begin{align*}
    L(\ell, \x)
    :&=
    \int_{\R^{m}}\Big\{ \Big(\ell(U_{\mfg(\t, \x)|\X}) - \gamma(U_{\mfg(\t, \x)|\X})\Big)^2
    \\
    &\qquad + \sum_{k = 1}^m \, 2\;\eta_k(\x)\,  t_k \, \partial_{k} \mfg(\t, \x) \ell(U_{\mfg(\t, \x)|\X}) 
      \Big\}\diff F_{\D|\X}(\t|\x)\  
      \\
      &= 
     \int_{\R^{m}}\Big\{ \Big(\ell(U_{\mfg(\t, \x)|\X}) - 
     \big[\gamma(U_{\mfg(\t, \x)|\X})
      - \sum_{k = 1}^m \;\eta_k(\x)\,  t_k \, \partial_{k} \mfg(\t, \x)\big]\Big)^2
      \Big\} \diff F_{\D|\X}(\t|\x) + c\,,
\end{align*}
where $c$ contains terms independent of $\ell$. Thus, the optimal $\ell^*$ satisfies for all $(\t, \x) \in \supp(\D, \X)$
\begin{equation*}
 \ell^*
      :=    \gamma(U_{\mfg(\t, \x)|\X})
      - \sum_{k = 1}^m \;\eta_k^*(\x)\,  t_k \, \partial_{k} \mfg(\t, \x)\,,
\end{equation*}
where the Lagrange multipliers are such that the constraints are fulfilled.
Finally, we calculate 
\begin{align*}
    \rho_{\ell^*}(Y \,|\,\X) 
       &=
      \rho_{\gamma}(Y \,|\,\X) 
           - 
           \sum_{k = 1}^m \;\eta_k^*(\X)
       \;\E[Y \, D_i \, \partial_{i} \mfg(\D, \X)\, |\,\X ]\,,
\end{align*}
which concludes the representation of the multi-marginal fair decision rule. Uniqueness, if the multi-marginal fair decision rule exists, follows by strict convexity of the objective function and linearity of the constraints.
\end{proof}

\begin{proof}[Proof of \Cref{prop:marginal-compact}]
    This follows similarly to the proof of \Cref{prop:marginal-sensitivity} using the chain rule and noting that $\P$-a.s.
    \begin{equation*}
        \lim_{\delta\to 0}\frac{F_{D_i}^{-1} \big(\Phi \big(\Phi^{-1}(U) (1  + \delta)\big) \big) - D_i}{\delta}
        = 
        \frac{\phi\big(\Phi^{-1}(U)\big)}{f_{D_i}\big(F_{D_i}^{-1} (U ) \big)}
        =
        \frac{\phi\big(\Phi^{-1}(F_{D_i}(D_i))\big)}{f_{D_i} (D_i )}\,,
    \end{equation*}
    which concludes the proof.
\end{proof}

\begin{proof}[Proof of \Cref{cor:marginal-compact}]
    The proof follows along the lines of the proofs of \Cref{thm:individual} and \Cref{prop:mutli-marginal} and is omitted.
\end{proof}

\begin{proof}[Proof of \Cref{prop:sens-discrete-mean}]
We apply in the third equation \Cref{lemma:delta-function-general,lemma:derivative-kappa}, 
\begin{align*}
    \partial_{D_i} \, \E[Y|\X]
    &=
    \lim_{\delta \to 0} \frac1\delta \, \E[Y_\delta - Y \, |\,\X  ]
    \\
    &= 
    \lim_{\delta \to 0} \frac1\delta \sum_{k = 1}^{K-1}\E\big[ \Delta_k \mfg\, \big(\Id_{\{ \Phi\left(\Phi^{-1}(\tilde{U}) (1+ \delta)\right)\le p_k\}} - \Id_{\{ \tilde{U} \le p_k\}}\big)  \, |\,\X \big]
    \\
    &= 
    \sum_{k = 1}^{K-1} v_k \, \E \big[\Delta_k \mfg \Id_{\{\tilde{U} = p_k\}} \, |\,\X \big]
    \\
    &=
    \sum_{k = 1}^{K-1} v_k \, \E \big[\Delta_k \mfg \Id_{\{D_i = t_k\}} \, |\,\X \big]\,,
\end{align*}
where the last equation follows by definition of $\tilde{U}$.
\end{proof}

\begin{proof}[Proof of \Cref{thm:sens-discrete}]
Similar to the proof of \Cref{prop:marginal-sensitivity}, we denote the distorted output by $Y_\delta:=\mfg\big( \D_{i,\delta} , \X\big) $, with perturbation given in \eqref{eq:D-discrete-pert}, and the conditional cdfs by $F(y):= \P(Y \le y |\X = \x, \D_{-i} = \d_{-i})$ and by $F(y, \delta):= \P(Y_\delta \le y |\X = \x, \D_{-i} = \d_{-i})$. For simplicity we write $U:= U_{Y|\X, \D_{-i}}$, then as $F^{-1}(U, \delta) $ and $Y_\delta$ have the same distribution, for all $\delta\ge 0$, we have
\begin{align*}
    \lim_{\delta \to 0}\tfrac{1}{\delta}\big(\rho_\gamma(Y_\delta  | \X) &- \rho_\gamma(Y | \X)\big)
    \\
    &=
    \E\Big[ \E\big[ \lim_{\delta \to 0}\tfrac{1}{\delta}(F^{-1}(U, \delta) - F^{-1}(U)) \gamma(U)\, \big|\, \X = \x, \D_{-i} = \d_{-i}\big] \, \big|\, \X = \x \Big]\,.
\end{align*}
Using the representation of $Y_\delta$ and $Y$ in \eqref{eq:Y-discrete-pert}, and denoting $\kappa_\delta(x):= \Phi\left(\Phi^{-1}(x) (1+ \delta)\right)$, the inner expectation becomes, 
\begin{align*}
    \E\big[\lim_{\delta \to 0}\tfrac{1}{\delta} (F^{-1}(U, \delta)& - F^{-1}(U)) \gamma(U)\, \big|\, \X = \x, \D_{-i} = \d_{-i}\big]
    \\
    &= 
    \sum_{k = 1}^{K-1}\Delta_k \mfg \, \E\big[ \lim_{\delta \to 0}\tfrac{1}{\delta}\big(\Id_{\{ \kappa_\delta(\tilde{U}) \le p_k\}}  - \Id_{\{ \tilde{U}\le p_k\}} \big) \gamma(U)\, \big|\, \X = \x, \D_{-i} = \d_{-i}\big] 
    \\
    &= 
     \sum_{k = 1}^{K-1}v_k\, \Delta_k \mfg \, \E[  \Id_{\{D_i = t_k\}} \,\gamma(U)\, | \X = \x, \D_{-i} = \d_{-i}]
    \end{align*}
where in the last equation we applied \Cref{lemma:delta-function-general,lemma:derivative-kappa}. Collecting, we obtain
    \begin{align*}
     \lim_{\delta \to 0}\tfrac{1}{\delta}\big(\rho_\gamma(Y_\delta  | \X) - \rho_\gamma(Y | \X)\big)
    =
    \sum_{k = 1}^{K-1}v_k\,  \E[\Delta_k \mfg \, \Id_{\{D_i = t_k\}} \,  \gamma(U_{Y|\X})\,| \X = \x]
\end{align*}
which concludes the proof.
\end{proof}

\begin{proof}[Proof of \Cref{cor:marginal-discrete}]
    The proof follows along the lines of \Cref{cor:marginal-compact} and is omitted.
\end{proof}

\begin{proof}[Proof of \Cref{thm:casaade-sensitivity}]
    The proof follows by applying \Cref{prop:marginal-sensitivity}. We also refer to \cite{Pesenti2021RA} who first proved the result under stronger assumptions.
\end{proof}

\begin{proof}[Proof of \Cref{prop:marginal-cascade}]
    The proof follows along the lines of the proof of \Cref{cor:marginal-compact} and is omitted.
\end{proof}

\begin{proof}[Proof of \Cref{prop:cascade-compact}]
    This follows from \Cref{prop:marginal-compact}.
\end{proof}

\begin{proof}[Proof of \Cref{prop:cascade-discrete}]
    First we note that
\begin{equation*}
    \partial_{D_i}^c\;  \rho_\gamma(\, Y\,|\,\X\,)
    =
    \sum_{l = 1}^{n+m} \lim_{\delta \to 0}\tfrac{1}{\delta}\Big(\rho_\gamma\big(\mfg\big((\D, \X)_{-l}, \Psi^{(l)}(D_{i, \delta}, \V)\big)\big) - \rho_\gamma(Y)\Big)\,,
\end{equation*}
where we use the notation that $(\D, \X)_{-l}$ is the vector $(\D, \X)$ deprived of the $l^\text{th}$ component, $l \in \{1, \ldots, m+n\}$. Then for each $l\in \{1, \ldots, m+n\}$, the perturbed output is
\begin{align*}
    \mfg\big((\D, \X)_{-l}, \Psi^{(l)}(D_{i, \delta}, \V)\big)
    =
    \sum_{k = 1}^K \Delta_{k,l} \tilde{\mfg}\, \Id_{\{ \Phi\left(\Phi^{-1}(\tilde{U}) (1+ \delta)\right)\le p_k\}}  + \tilde{\mfg}_{K,l}
        \,,
\end{align*}
where $ \Delta_{k,l} \tilde{\mfg}: =
\mfg\big((\D, \X)_{-l}, \Psi^{(l)}(t_k, \V)\big) - \mfg\big((\D, \X)_{-l}, \Psi^{(l)}(t_{k+1}, \V)\big)$, $k = 1, \ldots, K -1$, and $ \tilde{\mfg}_{K,l} := \mfg\big((\D, \X)_{-l}, \Psi^{(l)}(t_K, \V)\big)$. With this representation, we use for each $l \in \{1, \ldots, m+n\}$ similar steps as in the proof of \Cref{thm:sens-discrete} and obtain that
\begin{equation*}
    \partial_{D_i}^c\;  \rho_\gamma(\, Y\,|\,\X\,)
    =
    \sum_{l = 1}^{n+m}\, 
    \sum_{k = 1}^{K-1}
    \E[\Delta_{k,l} \tilde{\mfg} \, \Id_{\{D_i = t_k\}} \,  \gamma(U_{Y|\X})\,| \X = \x]\,.
\end{equation*}
Finally, noting that by the standard construction we can choose $\Psi^{(l)}(t_k, \V) = F_{D_l}^{-1}(V| D_i = t_k)$, for $l = 1, \ldots, m$, and $\Psi^{(l)}(t_k, \V) = F_{X_l}^{-1}(V| D_i = t_k)$, for $l = m+1, \ldots, m+n$, for a uniform rv $V$ independent of $D_i$, concludes the proof.
\end{proof}

\section{Model specifics of empirical implementation} \label{app:model}
This appendix collects additional information on the numerical implementation of \Cref{sec:empirical}. \Cref{tab:model_summary} collects the model configuration. LR means learning rate, HL indicates the number of hidden layers.

\begin{table}[ht]
\centering
\scriptsize
\resizebox{\textwidth}{!}{%
\begin{tabular}{l l l l l l p{4cm} l}
\toprule
\textbf{Task} & \textbf{Model Type} & \textbf{Loss Function} & \textbf{Input Features} & \textbf{Target} & \textbf{Optimizer} & \textbf{Hyperparameters} & \textbf{Additional Notes} \\
\midrule
Estimate $g()$ & GLM & Tweedie Loss & $X_{\text{train}}, D_{\text{train}}$ & $y_{\text{train}}$ & Adam & LR: 0.01 & log link \\
Estimate Expected Value $\rho()$ & GLM & Tweedie Loss & $X_{\text{train}}$ & $y_{\text{train}}$ & Adam & LR: 0.01 & log link \\
Estimate $\E\big[Y \, D_i \, \partial_{i} \mfg(\D, \X)\, |\,\X \big]$ & NN & Tweedie Loss & $X_{\text{train}}$ & $Y_{\text{train}} \times D_{\text{train}} \times \Delta_g$ & Adam & LR: 0.0001, HL: [100] & log link \\
Estimate $\E\big[\,\big(D_i \partial_{i} \mfg(\D, \X) \big)^2 \big|\,\X\big]$ & NN & Gamma Loss & $X_{\text{train}}$ & $(D_{\text{train}} \times \Delta_g)^2$ & Adam & LR: 0.0001, HL: [100] & log link \\
Predict $\Id_{\{D_i = t_k\}}$ using $\X$ & NN & Binary Cross Entropy & $X_{\text{train}}$ & $D_{\text{train}}$ & Adam & LR: 0.001, HL: [100] & -- \\
Quantile Regression to estimate VaR & Quantile Regression & pinball Loss & $X_{\text{train}}$ & $y_{\text{train}}$ & interior-point & $\alpha = 0.9$ & -- \\
Estimate ES $\rho()$ & GLM & Tweedie Loss & $X_{\text{train}} \geq VaR_{\alpha}$ & $y_{\text{train}}\geq VaR_{\alpha}$ & Adam & LR: 0.01, HL: [100] & log link \\
Predict $\Id_{\{D_i = t_k\}} \gamma(U_{Y|\X})$ using $\X$ for ES & NN & Binary Cross Entropy & $X_{\text{train}}$ & $D_{\text{train}}$ & Adam & LR: 0.0001, HL: [100] & --\\
\bottomrule
\end{tabular}
}
\caption{Summary of model configurations}
\label{tab:model_summary}
\end{table}


\end{appendix}

\bibliographystyle{siamplain}
\bibliography{references}

\end{document}